\documentclass{article}

\usepackage[preprint]{neurips_2024}

\usepackage[utf8]{inputenc} 
\usepackage[T1]{fontenc}    
\usepackage{hyperref}       
\usepackage{url}            
\usepackage{booktabs}       
\usepackage{amsfonts}       
\usepackage{nicefrac}       
\usepackage{microtype}      
\usepackage{xcolor}         

\usepackage{amsmath,amsfonts,bm}

\def\eqref#1{equation~\ref{#1}}

\def\1{\bm{1}}

\DeclareMathAlphabet{\mathsfit}{\encodingdefault}{\sfdefault}{m}{sl}
\SetMathAlphabet{\mathsfit}{bold}{\encodingdefault}{\sfdefault}{bx}{n}

\usepackage{graphicx}
\usepackage{array}
\usepackage{ifthen}
\usepackage[normalem]{ulem}
\usepackage{wrapfig}
\usepackage{adjustbox}
\usepackage{caption}
\usepackage{enumitem}
\usepackage{multirow}
\usepackage{changepage}
\usepackage[toc,page,header]{appendix}
\usepackage{minitoc}

\usepackage[capitalise]{cleveref}
\usepackage{diagbox}

\newcommand{\appref}[1]{Appendix~\ref{#1}}

\newcommand{\bx}{\mathbf{x}}

\newcommand{\bs}{\mathbf{s}}
\newcommand{\F}{\mathcal{F}}
\newcommand{\btheta}{{\boldsymbol{\theta}}}
\newcommand{\reals}{{\mathbb R}}
\newcommand{\norm}[1]{\left\|#1\right\|}

\newcommand{\cls}{{\sc [CLS]}}

\newcommand{\clstt}[1]{\texttt{\small #1}}

\newcommand{\niv}[2][]{%
    \ifthenelse{ \equal{#1}{} }
        {\textcolor{red}{(NH) #2}}
        {\textcolor{red}{(NH) \sout{#1} #2}}
}
\newcommand{\gilad}[2][]{%
    \ifthenelse{ \equal{#1}{} }
        {\textcolor{magenta}{(GY) #2}}
        {\textcolor{magenta}{(GY) \sout{#1} #2}}
}
\newcommand{\gal}[2][]{%
    \ifthenelse{ \equal{#1}{} }
        {\textcolor{green}{(GV) #2}}
        {\textcolor{green}{(GV) \sout{#1} #2}}
}
\newcommand{\yakir}[2][]{%
    \ifthenelse{ \equal{#1}{} }
        {\textcolor{purple}{(YO) #2}}
        {\textcolor{purple}{(YO) \sout{#1} #2}}
}

\title{\vspace{-6pt}Reconstructing Training Data From Real-World Models Trained with Transfer Learning\vspace{-6pt}}

\newcommand*{\affaddr}[1]{#1}
\newcommand*{\affmark}[1][*]{\textsuperscript{#1}}

\newcommand*{\asp}{\quad\quad}

\author{
\hspace{-16pt} Yakir Oz\affmark[1]\asp Gilad Yehudai\affmark[2]\asp Gal Vardi\affmark[1]\asp Itai Antebi\affmark[1]\asp Michal Irani\affmark[1]\asp Niv Haim\affmark[1] \\
\vspace{-4pt}\\
\affaddr{\affmark[1]Weizmann Institute of Science, Rehovot, Israel}\\
\affaddr{\affmark[2]Center for Data Science, New York University} 
}

\begin{document}

\doparttoc 
\faketableofcontents 

\maketitle
\vspace{-18pt}

\begin{abstract}
\vspace{-4pt}
Current methods for reconstructing training data from trained classifiers are restricted to very small models, limited training set sizes, and low-resolution images. Such restrictions hinder their applicability to real-world scenarios. In this paper, we present a novel approach enabling data reconstruction in realistic settings for models trained on high-resolution images. Our method adapts the reconstruction scheme of~\cite{haim2022reconstructing} to real-world scenarios -- specifically, targeting models trained via transfer learning over image embeddings of large pre-trained models like DINO-ViT and CLIP. Our work employs data reconstruction in the embedding space rather than in the image space, showcasing its applicability beyond visual data. Moreover, we introduce a novel clustering-based method to identify good reconstructions from thousands of candidates. This significantly improves on previous works that relied on knowledge of the training set to identify good reconstructed images. Our findings shed light on a potential privacy risk for data leakage from models trained using transfer learning. 
\end{abstract}

\vspace{-12pt}
\section{Introduction}
\vspace{-8pt}

Understanding when training data can be reconstructed from trained neural networks is an intriguing question that attracted significant interest in recent years. Successful reconstruction of training samples has been demonstrated for both generative models~\citep{carlini2021extracting,carlini2023extracting} and classification settings~\citep{haim2022reconstructing}. Exploring this question may help understand the extent to which neural networks memorize training data and their vulnerability to privacy attacks and data leakage.

Existing results on training data reconstruction from neural network classifiers focus on restricted and unrealistic settings. These methods require very small training datasets, which strongly limit their ability to generalize. Additionally, they are constrained to low-resolution images, such as CIFAR or MNIST images, and simple models like multilayered perceptrons (MLPs) or small CNNs.

We aim to overcome these limitations in a transfer-learning setting. Transfer Learning leverages knowledge gained from solving one problem to address a related problem, often by transferring learned representations from large pre-trained models (known as \emph{Foundation Models}) to tasks with limited training data. In the context of deep learning, transfer learning is commonly implemented by fine-tuning the final layers of pre-trained models or training small MLPs on their output embeddings, known as deep features~\citep{oquab2014learning}.
This approach often achieves
high generalization even for learning tasks with small training sets, while also requiring less computing power. Thus, transfer learning is very common in practice.

\begin{figure}[t]
    \centering
    \includegraphics[width=\textwidth]{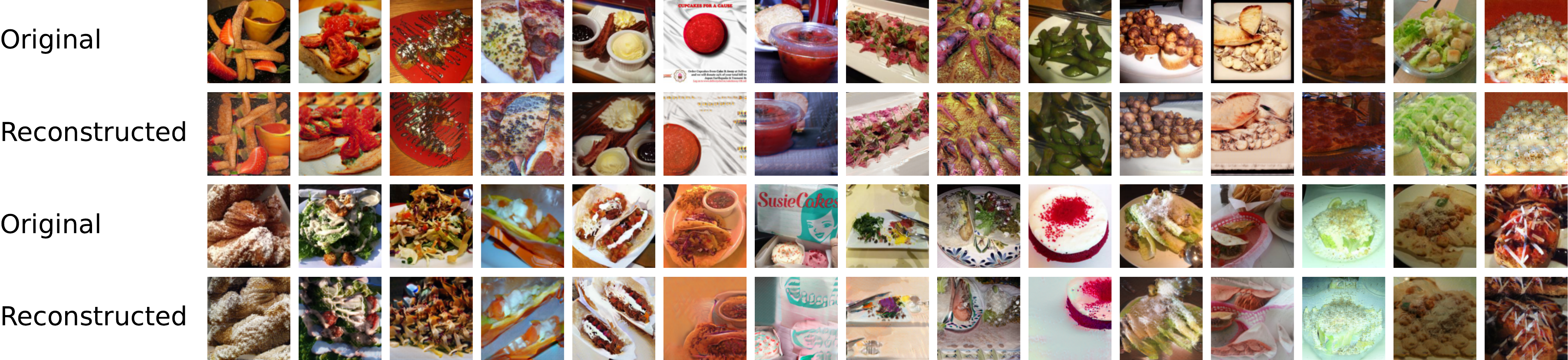}
    \caption{Reconstructed Data from a binary classifier trained on $100$ DINO-VIT embeddings\vspace{-12pt}}
    \label{fig:recon_dinovit}
\end{figure}

In this work, we demonstrate reconstruction of training samples in more realistic scenarios. Specifically, we reconstruct high-resolution images from models that achieve good test performance, within a transfer learning framework. Our approach involves training an MLP on the embeddings of common pre-trained transformer-based foundation models, such as CLIP~\citep{radford2021learning} or DINO-ViT~\citep{caron2021emerging} (see~\cref{fig:recon_dinovit}). Our findings have implications for privacy, particularly when transfer learning is being used on sensitive training data, such as medical data. Consequently, preventing data leakage in transfer learning necessitates the development of appropriate defenses.

Additionally, our work addresses a key limitation of prior reconstruction works: their reliance on training images for identifying good reconstructions from thousands of candidates.
While this approach demonstrated that training images are embedded within the model's parameters, it's unrealistic for attackers to have access to the training data. To overcome this, we introduce a novel clustering-based approach to effectively identify reconstructed training samples, eliminating the need for prior knowledge of the training set. This marks a significant step towards establishing reconstruction techniques as real-world privacy attacks.

\textbf{Our Contributions:}
\begin{itemize}
    \item We demonstrate reconstruction of high-resolution training images from models trained in a transfer learning approach, a significant advancement from previous reconstruction methods that were limited to small images and models with low generalization.
    \item We demonstrate, for the first time, reconstruction of non-visual data (feature vectors of intermediate layers).
    \item We introduce a novel clustering-based approach for effectively identifying training samples without a-priori knowledge of training images, a significant step towards a more realistic privacy attack.
\end{itemize}

\section{Prior Work}

\paragraph{Data Reconstruction Attacks.}

Reconstruction attacks attempt to recover the data samples on which a model is trained, posing a serious threat to privacy. Earlier examples of such attacks include activation maximization (model-inversion) \citep{fredrikson2015model,yang2019neural}, although they are limited to only a few training samples in each class. Reconstruction in a federated learning setup \citep{zhu2019deep,he2019model,hitaj2017deep,geiping2020inverting,huang2021evaluating,wen2022fishing} where the attacker assumes knowledge of samples' gradients. Other works studied reconstruction attacks on generative models like LLMs \citep{carlini2019secret,carlini2021extracting,nasr2023scalable} and diffusion-based image generators \citep{somepalli2022diffusion,carlini2023extracting}. Our work is based on the reconstruction method from \cite{haim2022reconstructing}, which relies only on knowledge of the parameters of the trained model, and is based on theoretical results of the implicit bias in neural networks \citep{lyu2019gradient,ji2020directional}. This work was generalized to multi-class setting \citep{buzaglo2023deconstructing} and to the NTK regime \citep{loo2023dataset}.

\paragraph{Transfer Learning.}
Deep transfer learning, a common technique across various tasks (see surveys: \citep{tan2018survey,zhuang2020comprehensive,iman2023review}), leverages pre-trained models from large datasets to address challenges faced by smaller, domain-specific datasets (e.g., in the medical domain~\citep{kim2022transfer}). While convolutional neural networks (CNNs) have been the go-to approach for transfer learning~\citep{oquab2014learning,yosinski2014transferable}, recent research suggests that vision transformers (ViTs) may offer stronger learned representations for downstream tasks~\citep{caron2021emerging,he2022masked}. For example, ViT~\citep{dosovitskiy2020image}, pre-trained on ImageNet~\citep{deng2009imagenet}, provides robust general visual features.
Beyond supervised pre-training, self-supervised learning methods like DINO~\citep{caron2021emerging,oquab2023dinov2} learn informative image representations without requiring labeled data, allowing the model to capture strong image features suitable for further downstream tasks. Additionally, CLIP~\citep{radford2021learning} has emerged as a powerful technique, leveraging a massive dataset of paired text-image examples and contrastive loss to learn semantically meaningful image representations.

\begin{figure*}[th]
    \centering
    \begin{tabular}{c}
        (a) Classifier trained via Transfer-Learning (on embedding-space) \\
         \includegraphics[width=.75\textwidth]{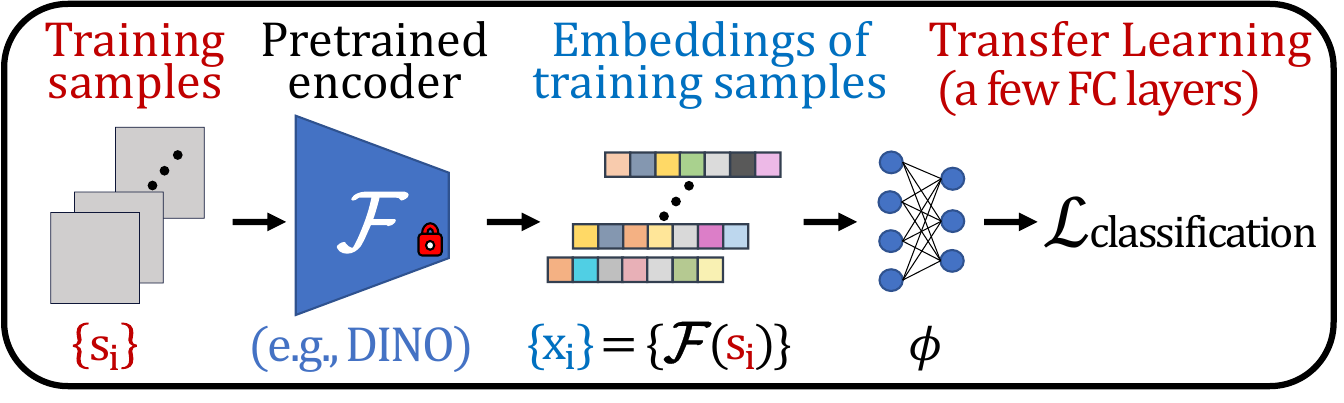}  \\
         (b) Training-Data Reconstruction from the Classifier \\
         \includegraphics[width=.75\textwidth]{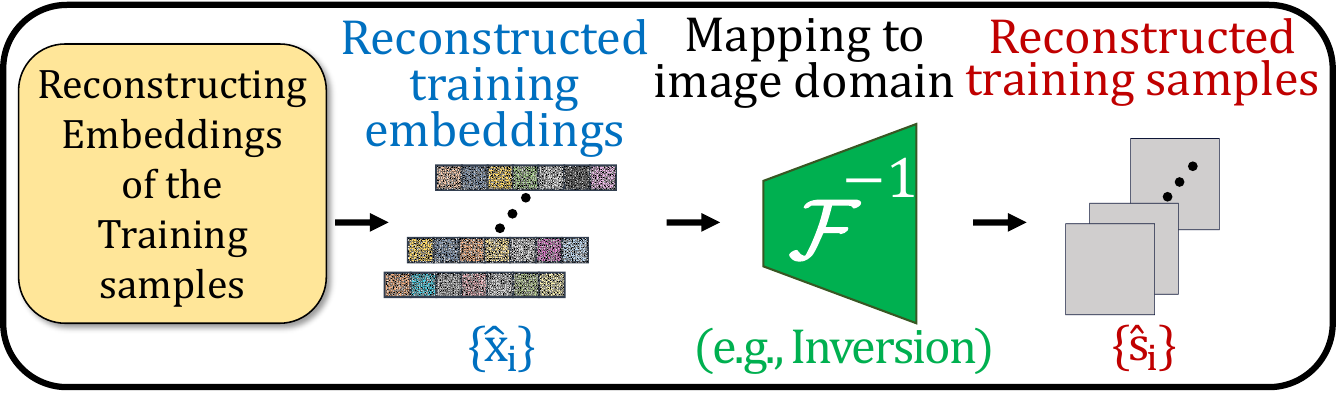}  \\
    \end{tabular}
    \caption{Overview of our training and data reconstruction scheme.}
    \label{fig:method_overview}
\end{figure*}

\section{Method}
\label{sec:Method}

Our goal is to reconstruct training samples (images) from a classifier that was trained on the corresponding embedding vectors of a large pre-trained model in a transfer learning manner.

The classifier training is illustrated in \cref{fig:method_overview}a. Formally, given an image classification task $D_s=\{(\bs_i, y_i)\}_{i=1}^n \subseteq \reals^{d_s} \times \{1,\ldots,C\}$, 
where $d_s$ is the dimension of the input image\footnote{Typically $d_s= {3 \times h \times w}$, where $h$ and $w$ are the height and width of the image, respectively.} and $C$ is the number of classes, we employ a large pre-trained model $\F: \reals^{d_s} \to \reals^d$ (e.g., DINO) to transfer each image $\bs_i$ to its corresponding deep feature embedding $\bx_i=\F(\bs_i)\in \reals^d$, where $d$ is the dimension of the feature embedding vector (the output of $\F$). We then train a model \mbox{$\phi(\cdot, \btheta): \reals^{d} \to \reals^C$} to classify the embedding dataset $D_x=(\bx_i, y_i)_{i=1}^n \subseteq \reals^{d} \times \{1,\ldots,C\}$, where $\btheta\in\reals^p$ is a vectorization of the trained parameters. Typically, $\phi$ is a single hidden-layer multilayer perceptron (MLP). Also note that $\F$ is kept fixed during the training of $\phi$. The overall trained image classifier is $\phi(\F(\bs))$.

Our reconstruction approach is illustrated in~\cref{fig:method_overview}b and presented in detail below. Given the trained classifier $\phi$ and the pre-trained model $\F$, our goal is to reconstruct training samples $\bs_i$ from the training set $D_s$. The reconstruction scheme comprises two parts:
\begin{enumerate}[leftmargin=.7cm]
    \item Reconstructing embedding vectors from the training set of the classifier $\phi$.
    \item Mapping the reconstructed embedding vectors back into the image domain. Namely, computing $\F^{-1}$ (e.g., by ``inverting'' the pre-trained model $\F$).
\end{enumerate}

\subsection{Reconstructing Embedding Vectors from \texorpdfstring{$\phi$}{}}
\label{sec:method_reconstruction}

Given a classifier $\phi: \reals^{d} \to \reals^c$ trained on an embedding training-set $D_x=\{(\bx_i, y_i)\}_{i=1}^n$, we apply the reconstruction scheme of~\citep{haim2022reconstructing,buzaglo2023deconstructing} to obtain $\{\hat{\bx}_i\}_{i=1}^m$, which are $m$ ``candidates'' for reconstructed samples from the original training set $D_x$. In this section we provide a brief overview of the reconstruction scheme of~\citep{haim2022reconstructing,buzaglo2023deconstructing} (for elaboration see Sec.~3~in~\cite{haim2022reconstructing}):

\paragraph{Implicit Bias of Gradient Flow:} \cite{lyu2019gradient,ji2020directional} show that given a homogeneous\footnote{W.r.t the parameters $\btheta$. Namely $\forall c>0: \phi(\cdot,c \btheta)=c^L \phi(\cdot,\btheta)$ for some $L$.} neural network $\phi(\cdot,\btheta)$ trained using gradient flow with a binary cross-entropy loss on a binary classification dataset $\{(\bx_i,y_i)\}_{i=1}^n\subseteq\reals^d\times \{\pm 1\}$, its parameters $\btheta$ converge to a KKT point of the maximum margin problem. In particular, there exist $\lambda_i \geq 0$ for every $i\in[n]$ such that the parameters of the trained network $\btheta$ satisfy the following equation:

\begin{equation}
    \label{eq:kkt}
    \btheta = \sum_{i=1}^n\lambda_i y_i\nabla_\btheta (\phi(\bx_i,\btheta))~.    
\end{equation}

\paragraph{Data Reconstruction Scheme:} Given such a trained model $\phi$ with trained (and fixed) parameters $\btheta$, the crux of the reconstruction scheme is to find a set of $\{\bx_i, \lambda_i, y_i\}$ that satisfy~\cref{eq:kkt}. This is done by minimizing the following loss function:
\begin{equation}
    \label{eq:reconstruction_loss}
    L_{\text{rec}}(\hat{\bx}_1,\dots,\hat{\bx}_m,\lambda_1,\dots,\lambda_m):= \norm{\btheta - \sum_{i=1}^m\lambda_i y_i\nabla_\btheta (\phi(\hat{\bx}_i,{\btheta}))}_2^2~,
\end{equation}

Where the optimization variables $\{\hat{\bx}_i, \lambda_i\}$ are initialized at random from $\lambda_i \sim \mathcal{U}(0,1)$ and $\hat{\bx}_i \sim \mathcal{N}(0, \sigma)$ ($\sigma$ is a hyperparameter). This generates \(m\) vectors $\{\hat{\bx}_i\}_{i=1}^m$ that we consider as ``candidates'' for reconstructed samples from the original training set of the classifier $\phi$. The number of candidates $m$ should be ``large enough'' (e.g., $m\geq2n$, and see discussion in~\cite{haim2022reconstructing}). The $y_i$ are assigned in a balanced manner (i.e., $y_1,\ldots,y_{m/2}=1$ and $y_{1+m/2},\ldots,y_{m}=-1$). Lastly, \cite{buzaglo2023deconstructing} extended this scheme to multi-class classification problems. 

The data reconstruction scheme is conducted multiple times for different choices of hyperparameters (e.g., learning rate and $\sigma$). For each trained model, we run about $50$-$100$ reconstruction runs with $m=500$, resulting in about $25$k-$50$k candidates. See \cref{appen:sweeps_details} for full details.

\subsection{Mapping Embedding Vectors \texorpdfstring{$\hat{\bx}_i$}{} to the Image Domain \texorpdfstring{$\hat{\bs}_i$}{}}
\label{sec:method_inversion}

Unlike previous works on data reconstruction that directly reconstruct training images, our method reconstructs embedding vectors. To evaluate the effectiveness of our reconstructed candidates, we must first map them back to the image domain. In this section we describe how we achieve training {\it images} from image-{\it embeddings}. Namely, given reconstructed image-embeddings $\hat{\bx}_i$, our goal is to compute $\hat{\bs}_i=\F^{-1}(\hat{\bx}_i)$. To this end we apply model-inversion methods and in particular, the method proposed in~\cite{tumanyan2022splicing}.

Given a vector $\hat{\bx}_i$ (an output candidate from the reconstruction optimization in~\cref{sec:method_reconstruction}), we search for an input image $\hat{\bs}_i$ to $\F$ that maximizes the cosine-similarity between $\F(\hat{\bs}_i)$ and $\hat{\bx_i}$. Formally:

\begin{equation}
    \label{eq:inversion_cossim}
    \hat{\bs}_i = \F^{-1}(\hat{\bx}_i) = \underset{\nu}{\text{argmax}} \frac{\F(\nu) \cdot \hat{\bx}_i}{\Vert\F(\nu)\Vert \Vert\hat{\bx}_i\Vert}~.
\end{equation}

We further apply a Deep-Image Prior (DIP)~\citep{ulyanov2018deep} to the input of $\F$. I.e., $\nu=g(\textbf{z})$ where $g$ is a CNN U-Net model applied to a random input $\textbf{z}$ sampled from Gaussian distribution. The only optimization variables of the inversion method are the parameters of $g$. See  \appref{appen:imp_inversion} further explanation and full implementation details.

By applying model-inversion to DINO embeddings,
\citet{tumanyan2022splicing} demonstrated that the \cls~ token contains a significant amount of information about the visual appearance of the original image from which it was computed. Even though their work was done in the context of image to image style transfer, their results inspired our work and motivated us to apply their approach in the context of reconstructing training image samples.

 A significant modification to \cite{tumanyan2022splicing} in our work is by employing a cosine-similarity loss instead of their proposed MSE loss. We find that using MSE loss (i.e., $\F^{-1}(\hat{\bx})={\text{argmin}}_{\nu}\Vert \F(\nu) - \hat{\bx}_i \Vert^2 $) is highly sensitive to even small changes in the scale of $\hat{\bx}$. The scales of $\hat{\bx}$ can be very different from the unknown $\bx=\F(\bs)$. Using cosine similarity alleviates this issue while simultaneously achieving similar quality for the inverted image result (see also~\cref{appen:cossim_motivation}).

The above-mentioned technique is used for mapping embeddings to images for most transformers that we consider in our work. However, this technique did not produce good results when applied to CLIP~\citep{radford2021learning}. Therefore, to map CLIP image embeddings to the image domain, we employ a diffusion-based generator conditioned on CLIP embeddings by~\cite{kakaobrain2022karlo-v1-alpha} (similar in spirit to the more popular DALL-E2~\citep{ramesh2022hierarchical}; see also~\cref{appen:clip_dip_inversion} and \cref{appen:unclip_inversion}).

\subsection{Selecting Reconstructed Embeddings to be Inverted} 
\label{sec:method_selection}

Applying the model-inversion described in~\cref{sec:method_inversion} to a large pretrained model is computationally intensive. Inverting a single embedding vector takes about $30$ minutes on an {\sc NVIDIA-V100-32GB GPU}. Therefore, it is not feasible to invert all $25$k-$50$k output candidates of~\cref{sec:method_reconstruction}. 

To determine which reconstructed candidates to invert, we pair each training embedding $\bx_i$ with its nearest reconstructed candidate $\hat{\bx}_j$ (measured by cosine similarity) and select the top $40$ vectors with the highest similarity for inversion. This approach proves effective in practice, yielding images with high visual similarity to the original training images, as demonstrated in the results (e.g.,~\cref{fig:recon_dinovit}).

In practice, the original training embeddings are not available (and inverting all candidates is computationally prohibitive). In~\cref{sec:clustering} we introduce a novel method to identify good reconstructions without relying on either ground-truth embeddings or exhaustive inversion.

\section{Results}

We demonstrate reconstructed training images from models trained in a transfer learning setup, on the embeddings of large pretrained models. We train several MLPs to solve learning tasks for various choices of training images and choices of the large pretrained backbones from which the image embeddings are computed.

\paragraph{Datasets.} Since we simulate a model that is trained in a transfer learning manner, it is reasonable to assume that such tasks involve images that were not necessarily included in the training sets on which the pretrained backbone was trained (typically, ImageNet~\citep{deng2009imagenet}). In our experiments we use images from \textbf{Food-101}~\citep{bossard2014food} (most popular dishes from foodspotting website) and \textbf{iNaturalist}~\citep{van2018inaturalist} (various animals/plants species) datasets. The resolution of the images vary between 250-500 pixels, but resized and center-cropped to $224\times224$. 

 
\paragraph{Pretrained Backbones ($\F$) for Image Embeddings.}
We select several Transformer-based foundation models that are popular choices for transfer learning in the visual domain:
\begin{itemize}[leftmargin=0.5cm]
    \item \textbf{ViT}~\citep{dosovitskiy2020image}: vit-base-patch16-224 from TIMM~\cite{rw2019timm}.
    \item \textbf{DINO-ViT}~\citep{caron2021emerging}: dino-vitb16 from the official implementation\footnote{\url{https://github.com/facebookresearch/dino}}.
    \item \textbf{DINOv2}~\citep{oquab2023dinov2}: dinov2-vitb14-reg from the official implementation\footnote{\url{https://github.com/facebookresearch/dinov2}}.
    \item \textbf{CLIP-ViT}~\citep{radford2021learning}: ViT-L/14 as provided by OpenAI’s CLIP repository \footnote{\url{https://github.com/openai/CLIP}}.
\end{itemize}

The dimension of the output embeddings is consistent across all backbones $\F$, and equal to $d$=$768$.

\paragraph{Multilayer Perceptron ($\phi$)} consists of a single hidden layer of dimension $500$ ($d$-$500$-$C$) that is optimized with gradient descent for $10$k epochs, weight-decay of $0.08$ or $0.16$ and learning rate $0.01$. All models achieve zero training error.

\subsection*{Reconstructing Training Data from \texorpdfstring{$\phi(\F)$}{}}

\newcolumntype{C}{ >{\centering\arraybackslash} m{2cm} }
\newcolumntype{D}{ >{\centering\arraybackslash} m{6cm} }
\begin{figure*}
    \centering
    \hspace*{-2cm}\begin{tabular}{C D D}
    \centering
    \backslashbox{\raisebox{-20pt}{\kern-1.6em\shortstack{Pretrained\\Model ($\F$)}}}{\kern-1emDataset} & Food101 & iNaturalist \\
    ViT & \includegraphics[width=.45\textwidth]{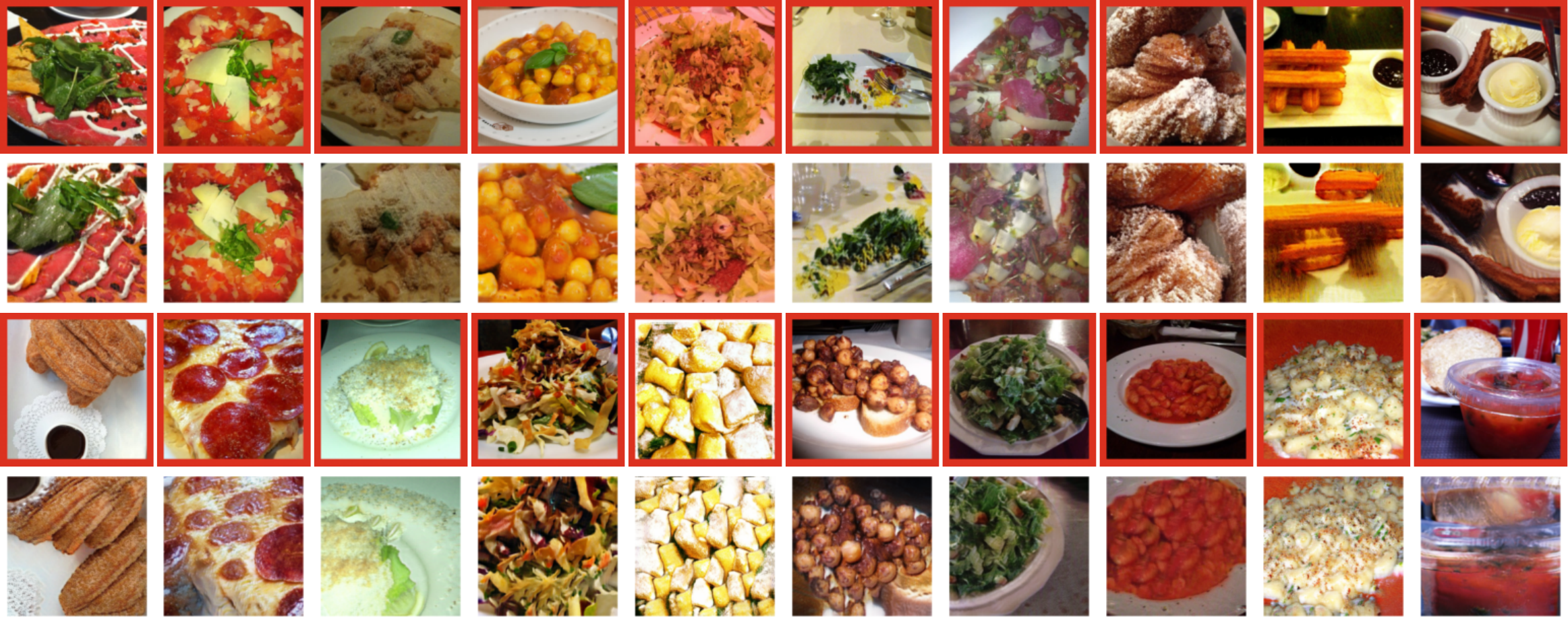} & \includegraphics[width=.45\textwidth]{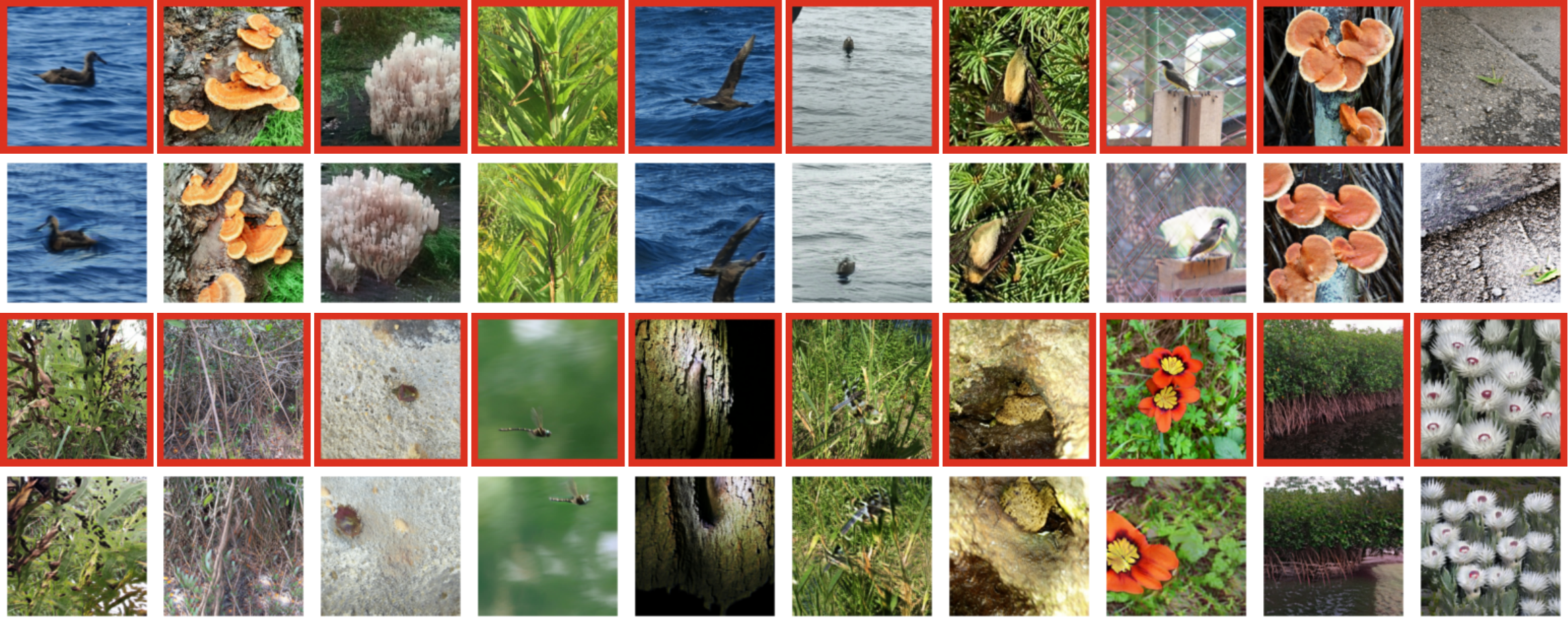} \\
    DINO-ViT & \includegraphics[width=.45\textwidth]{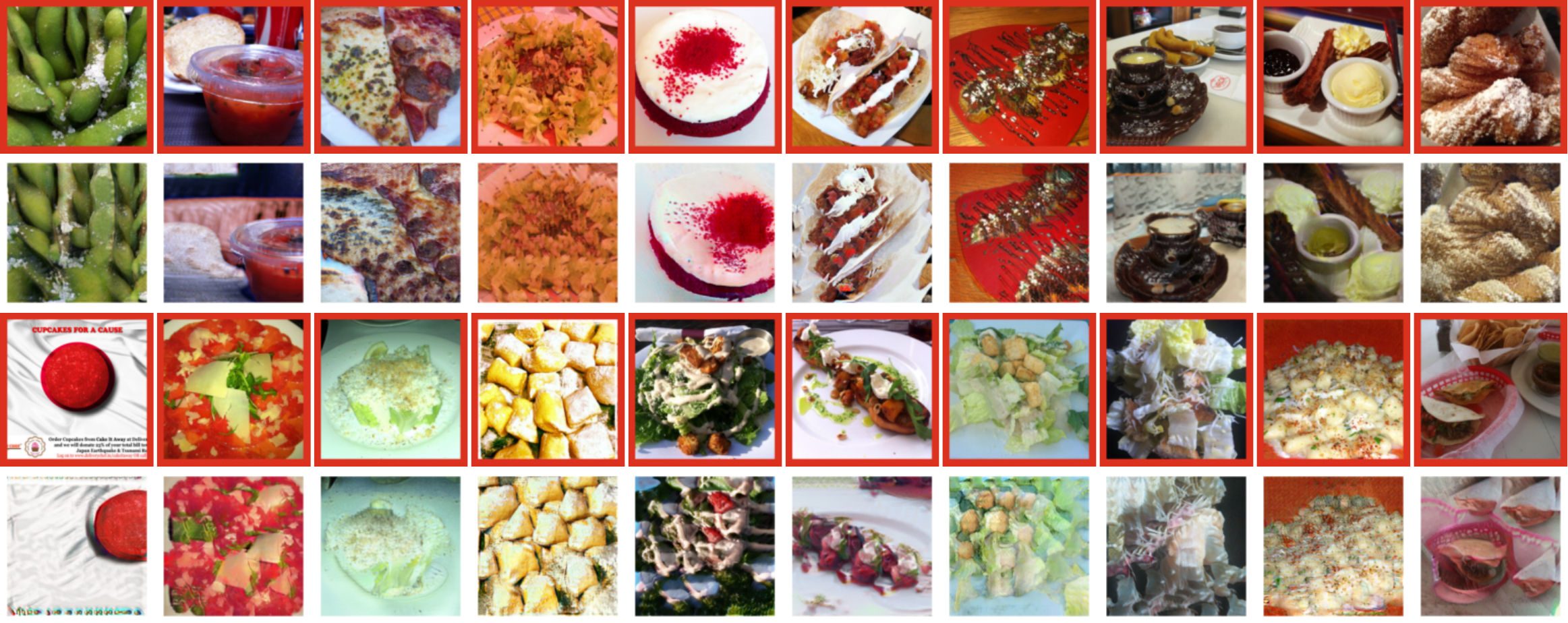} & \includegraphics[width=.45\textwidth]{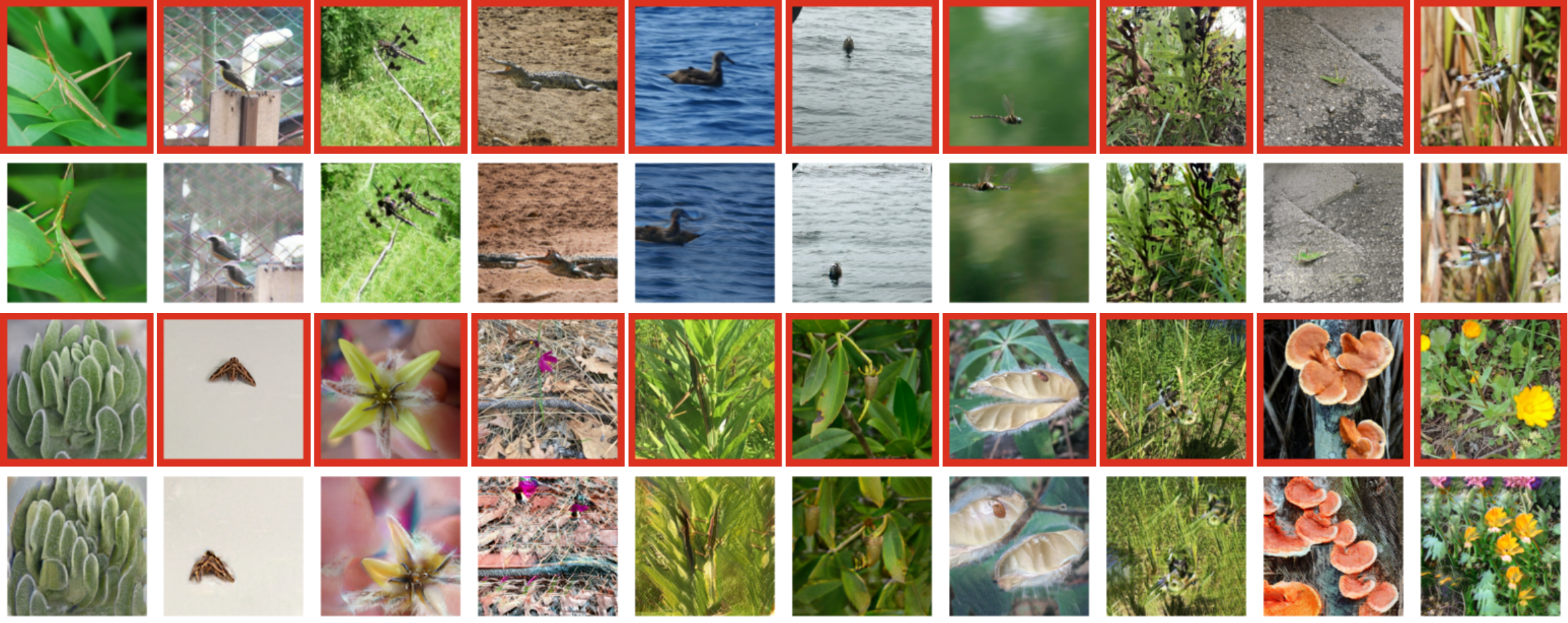} \\
    DINOv2-ViT & \includegraphics[width=.45\textwidth]{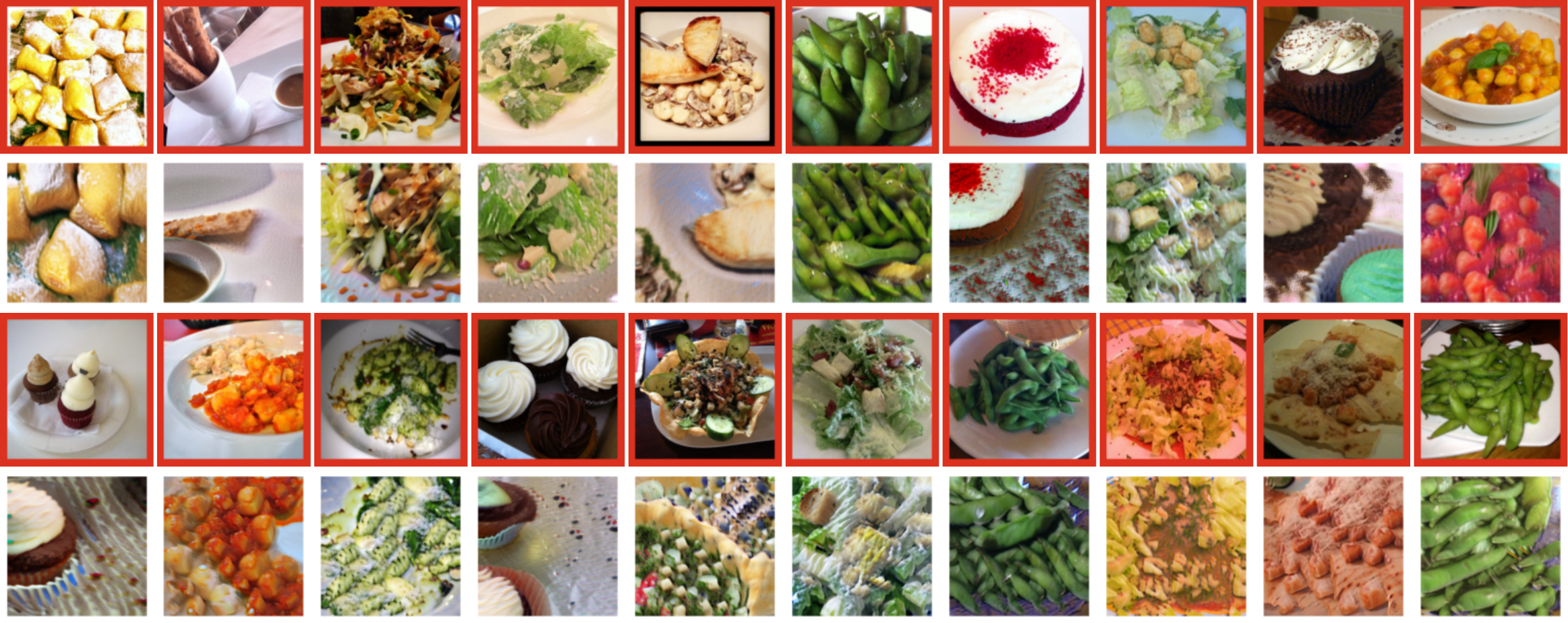} & \includegraphics[width=.45\textwidth]{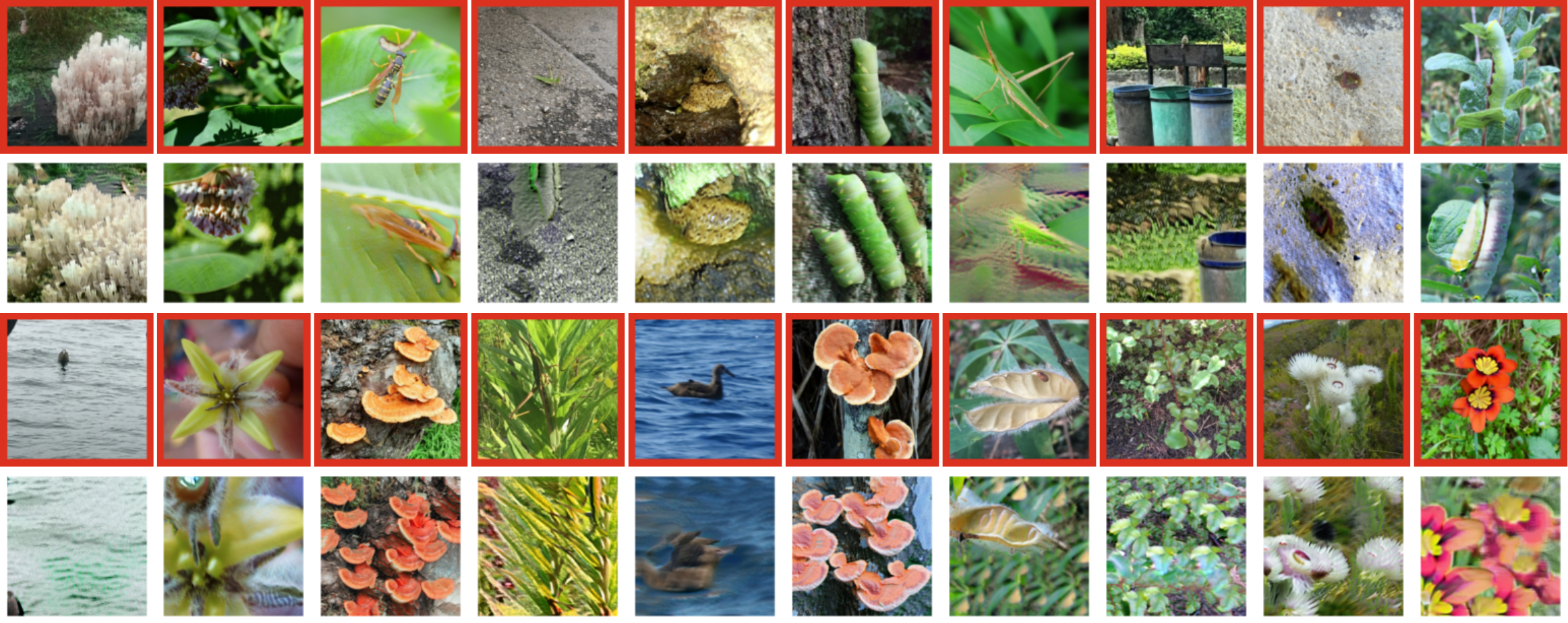} \\
    CLIP & \includegraphics[width=.45\textwidth]{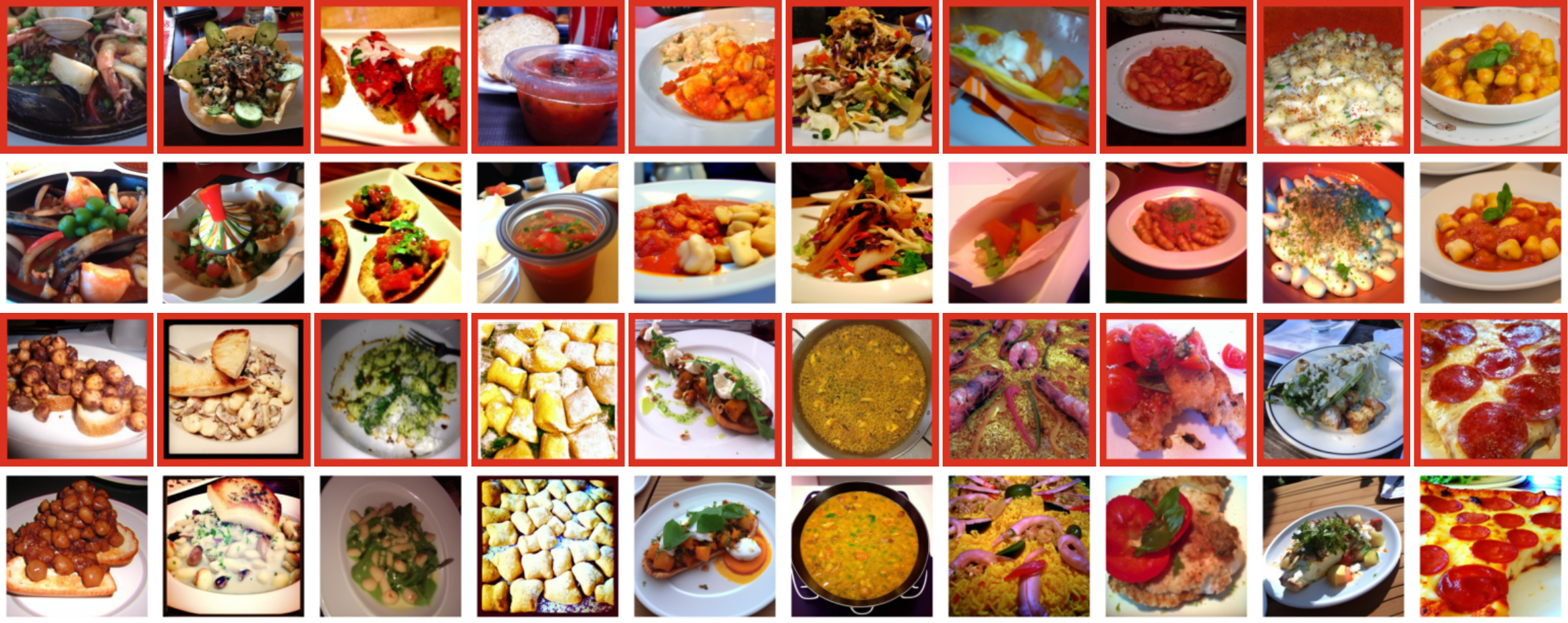} & \includegraphics[width=.45\textwidth]{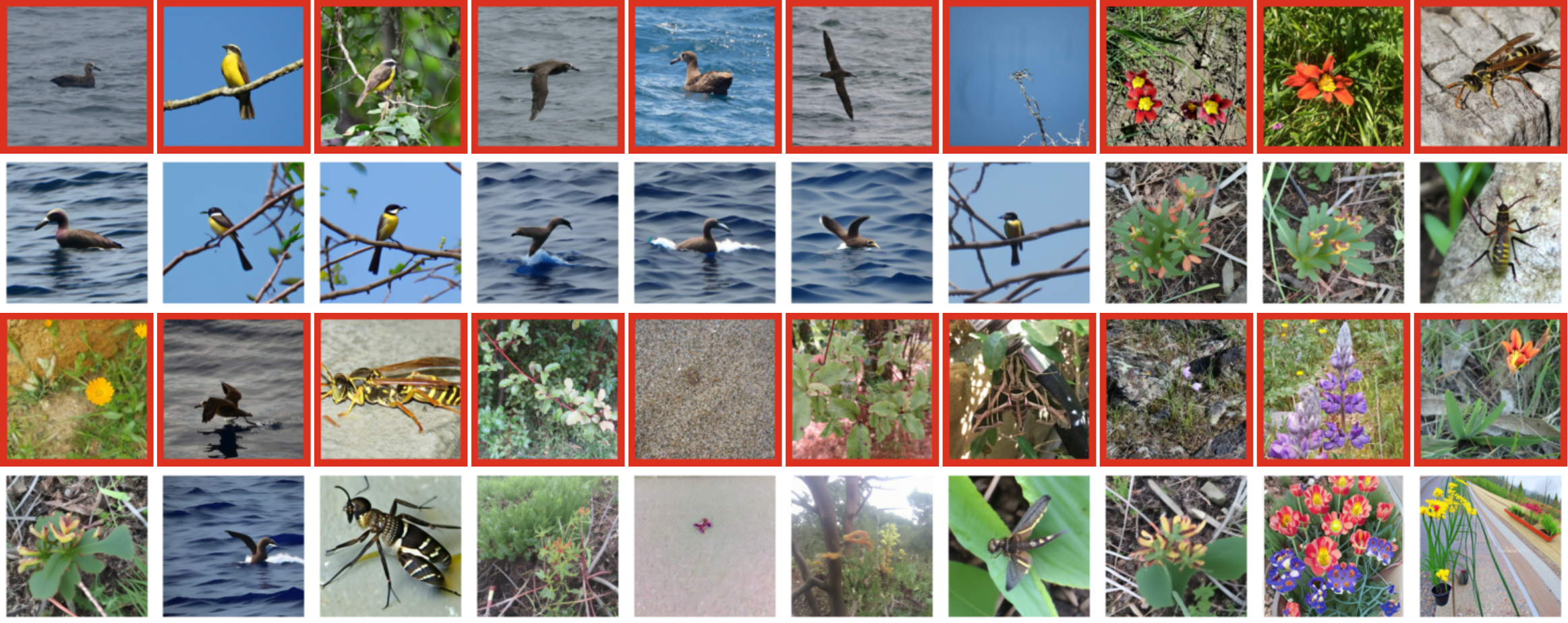} \\
    \end{tabular}
    \caption{Training samples (red) and their best reconstructed candidate, from MLPs trained on embeddings of various backbone models for two datasets.
    }
    \label{fig:main_results}
\end{figure*}

We train classifiers $\phi(\F(s))$ on two binary classification tasks: 
(1) {\it binary iNaturalist} is fauna (bugs/snails/birds/alligators) vs. flora (fungi/flowers/trees/bush) and (2) {\it binary Food101} is beef-carpaccio/bruschetta/caesar-salad/churros/cup-cakes vs. edamame/gnocchi/paella/pizza/tacos.
Each binary class mixes images from several classes of the original dataset (images are not mixed between different datasets). Each training set contains $100$ images ($50$ per class). The test sets contains $1000$/$1687$ images for iNaturalist/Food101 respectively. All models achieve test-accuracy above $95\%$ (except for DINO on Food101 with $85\%$ and see~\cref{fig:wd_selection}).

In~\cref{fig:main_results} we show the results of reconstructing training samples from $8$ models (for two binary tasks and $4$ choices of $\F$). For each reconstructed image ($\hat{\bs}=\F^{-1}(\hat{\bx})$), we show the nearest image from the training set, in terms of cosine-similarity between the embeddings of both ($d_{cosine}(\hat{\bx}, \F(\bs))$). The samples are ordered by their similarity (top-left to bottom-right).
As can be seen, many reconstructed images clearly have high semantic similarity to their corresponding nearest training images.

\begin{figure}[t]
    \centering
    \hspace{-1.5cm}
    \begin{tabular}{cc}
         \includegraphics[width=.46\textwidth]{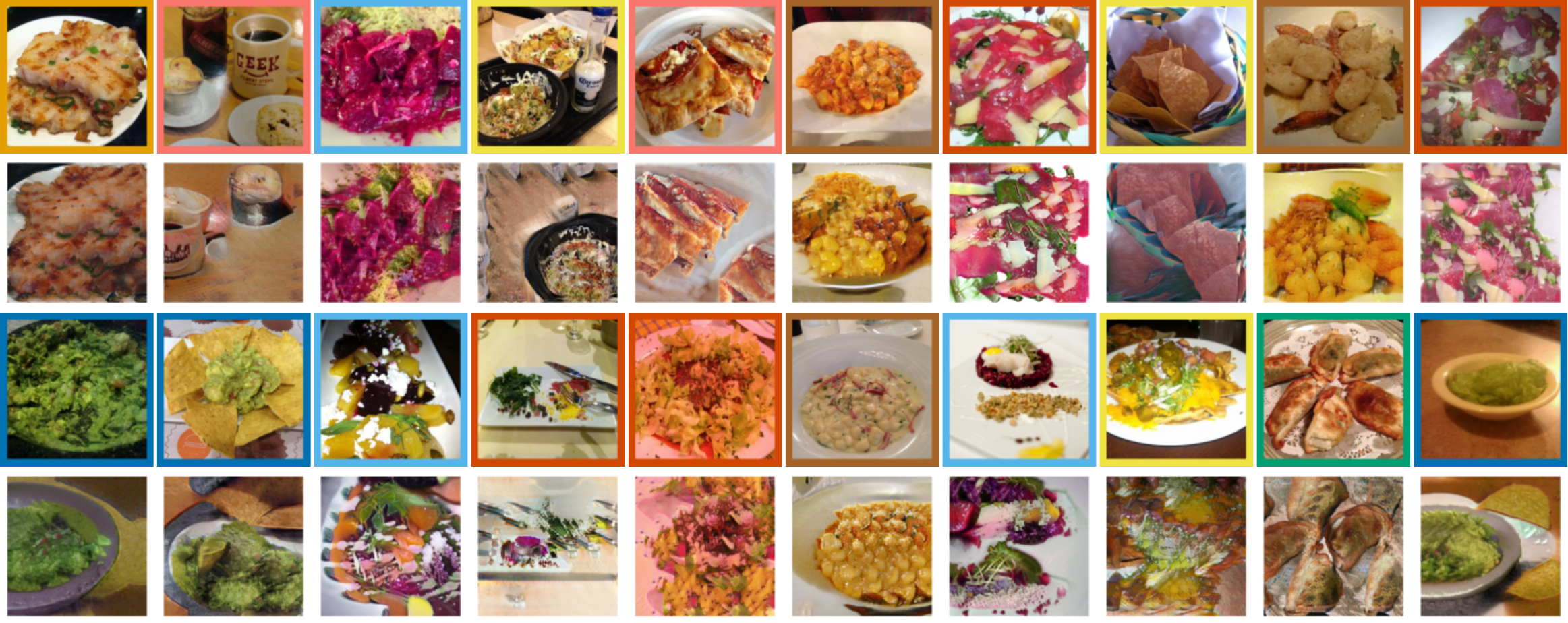} &
         \includegraphics[width=.46\textwidth]{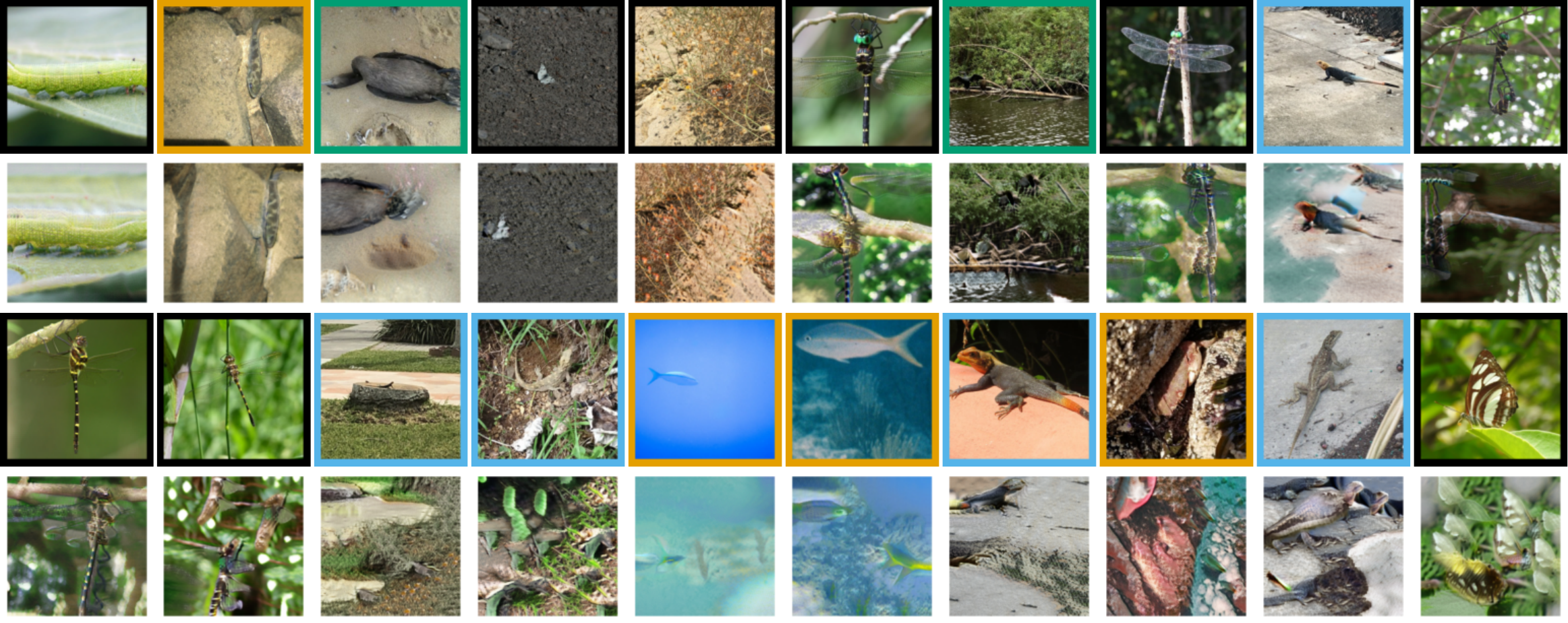} \\
         (a) DINO Food101 $10$ classes & (b) DINO iNaturalist $4$ classes \\
    \end{tabular}
    \caption{Reconstructions from a multiclass models trained $100$ images from Food101/iNaturalist with $C$=$10/4$ classes ($10/25$ images per class), with test-accuracy $84\%$/$96\%$ (on a/b respectively). Color-padded images are training images, where color represents different classes.}
    \label{fig:multiclass_10}
\end{figure}

In~\cref{fig:multiclass_10} we show results for reconstructing training samples from models trained on multiclass tasks, using an extension to of the method described in~\cref{sec:method_reconstruction}, to multiclass (by~\cite{buzaglo2023deconstructing}) and see also~\cref{appen:multiclass}. As can be seen our approach also extends to the multiclass setting.

The resemblance between images decreases as embedding similarity decreases. However, there's often still a visible semantic resemblance between the reconstructed and training images. One limitation of our reconstruction method is the varying effectiveness of the inversion method across different backbones $\F$. DINO and ViT yield the highest quality reconstructed samples. DINOv2 proves harder to invert, resulting in lower reconstruction quality. With CLIP, we utilize UnCLIP\footnote{We use the UnCLIP implementation from \url{https://github.com/kakaobrain/karlo}} to project embeddings into good natural images, maintaining semantic similarity even as reconstruction quality decreases (e.g., same class). See discussion in~\cref{sec:limitations} for further details on the differences and limitations of inversion.

\begin{figure}[htbp]
    \centering
    \begin{tabular}{c}
         \includegraphics[width=\textwidth]{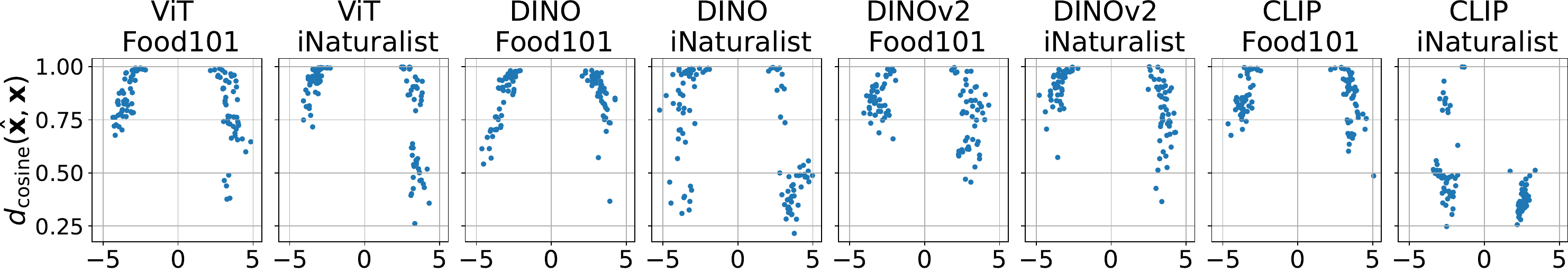}  \\
         \includegraphics[width=\textwidth]{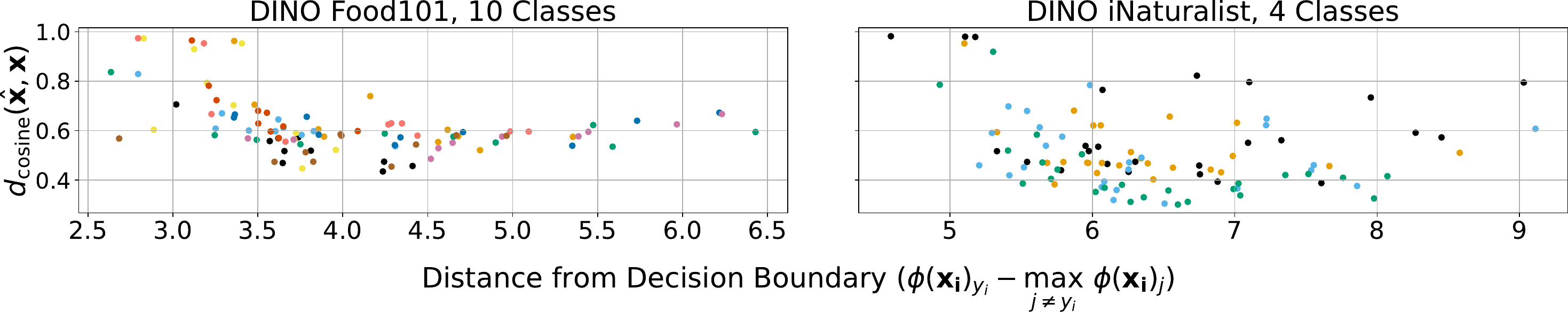} 
    \end{tabular}
    \caption{
    Numerical summary for all models whose reconstructed samples are shown in~\cref{fig:main_results,fig:multiclass_10}.
    } 
    \label{fig:cossim_preds}
\end{figure}

\cref{fig:cossim_preds} shows each training sample's reconstruction quality, measured by $d_{cosine}(\hat{\bx}, \F(\bs))$, plotted against its proximity to the decision boundary, measured by the model's output. There is a strong correlation between these quantities, aligning with the theory (Sec. 3.2/5.3 in~\cite{haim2022reconstructing}): if the classifier's parameters converge to the KKT solution of the maximum-margin problem, they only rely on data points on the classification margin (closest to the decision boundary). Such plots effectively summarize the reconstruction method's results, and in our case, also show its success.

\section{Identifying Good Reconstruction Without the Original Trainset}
\label{sec:clustering}

\begin{figure}
    \centering
    \includegraphics[width=\textwidth]{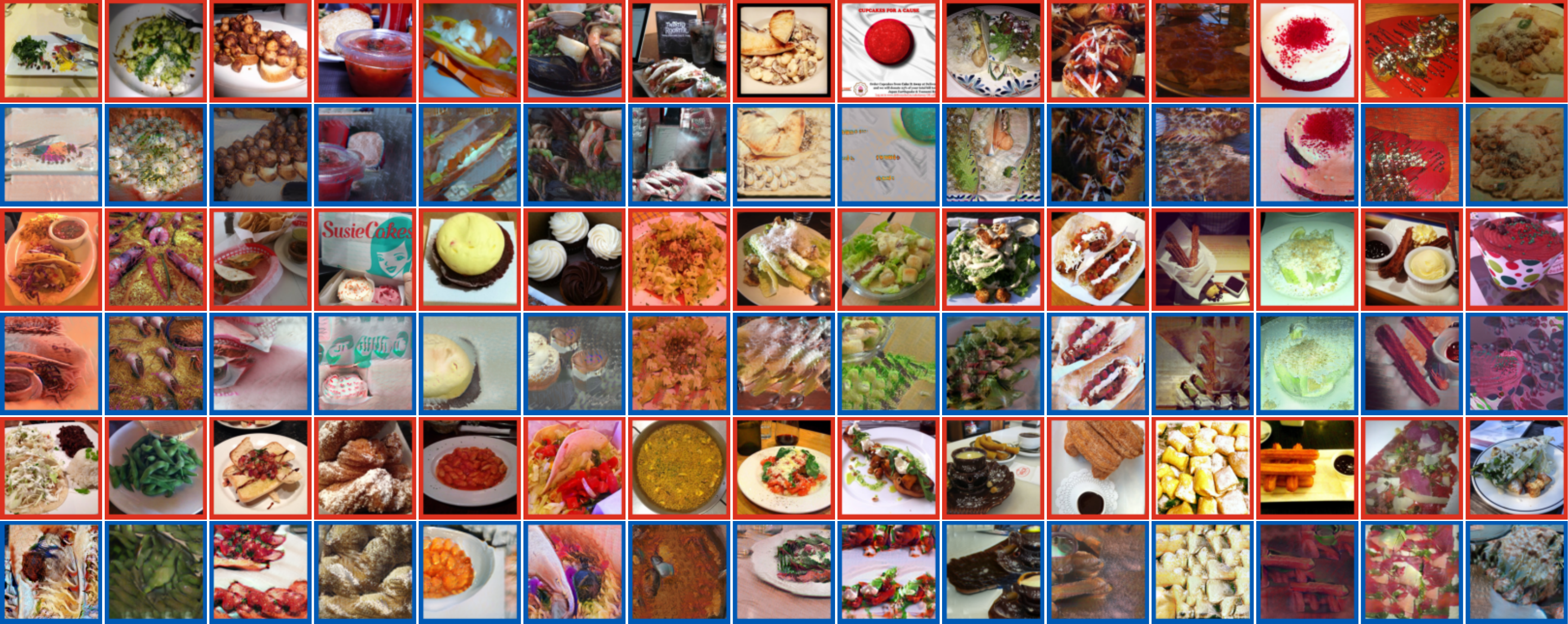}
    \caption{\textbf{Clustering-Based Reconstruction.} Inversion of clusters representatives (blue) compared to training samples whose embeddings are in the same cluster (in red).}
    \label{fig:clustering}
\end{figure}

In this section, we introduce a clustering-based approach to identify ``good''\footnote{Determining that a reconstructed candidate is indeed the reconstruction of a specific training image is equivalent to finding a good visual metric between images. See discussion in Sec.3 in~\cite{buzaglo2023deconstructing}.} reconstructed candidates without relying on the original training data. This is an important step towards an effective privacy attack. Previous works~\citep{haim2022reconstructing,buzaglo2023deconstructing,loo2023dataset}, including~\cref{sec:method_selection} in this work, rely on the original training images for demonstrating that training images are embedded in the model parameters. However, it is not applicable to real-world privacy attacks, as attackers don't have access to the original training data.

When directly reconstructing training images, this issue can be mitigated by manual inspection of the thousands of output image candidates – a time-consuming but feasible approach. However, this approach is irrelevant when reconstructing image embeddings. The reconstructed embeddings must first be inverted into images, which is computationally expensive (inverting a single vector takes about 30 minutes on an NVIDIA-V100-32GB GPU, as detailed in~\cref{sec:method_selection}). Inverting thousands of embeddings is simply infeasible.

This is where our proposed clustering approach comes in. We observe that reconstructed candidates whose inversions are visually similar to training samples tend to cluster together.
By applying clustering algorithms, we group similar candidates and only invert representative samples from the largest clusters. 
This reduces the total number of inversions by two orders of magnitude (from thousands to tens) and eliminates reliance on training data for identifying good reconstructed samples.

We demonstrate this by using agglomerative clustering\footnote{\scriptsize \url{https://docs.scipy.org/doc/scipy/reference/generated/scipy.cluster.hierarchy.fcluster.html}} on $25$,$000$ candidates reconstructed from a Dino-ViT-based model trained on the Food101 dataset (same as in~\cref{fig:main_results}). We use cosine similarity as the distance metric with ``average'' linkage and $1$,$000$ clusters, from which we select the 45 largest ones (containing between 100 and 8,000 candidates each). Within each cluster, a representative is chosen by averaging all candidate embeddings. Finally, these representatives are inverted using the methods described in~\cref{sec:method_inversion}.~\cref{fig:clustering} shows the results of inverting these cluster representatives (blue), along with a training sample whose embedding belongs to the same cluster (red). As can be seen, the clustering-based approach provides a very good method for reconstructing training samples without requiring the  training data.

\begin{wrapfigure}[15]{r}{0.35\textwidth}
    \centering
    \vspace{-10pt}
    \includegraphics[width=0.35\textwidth]{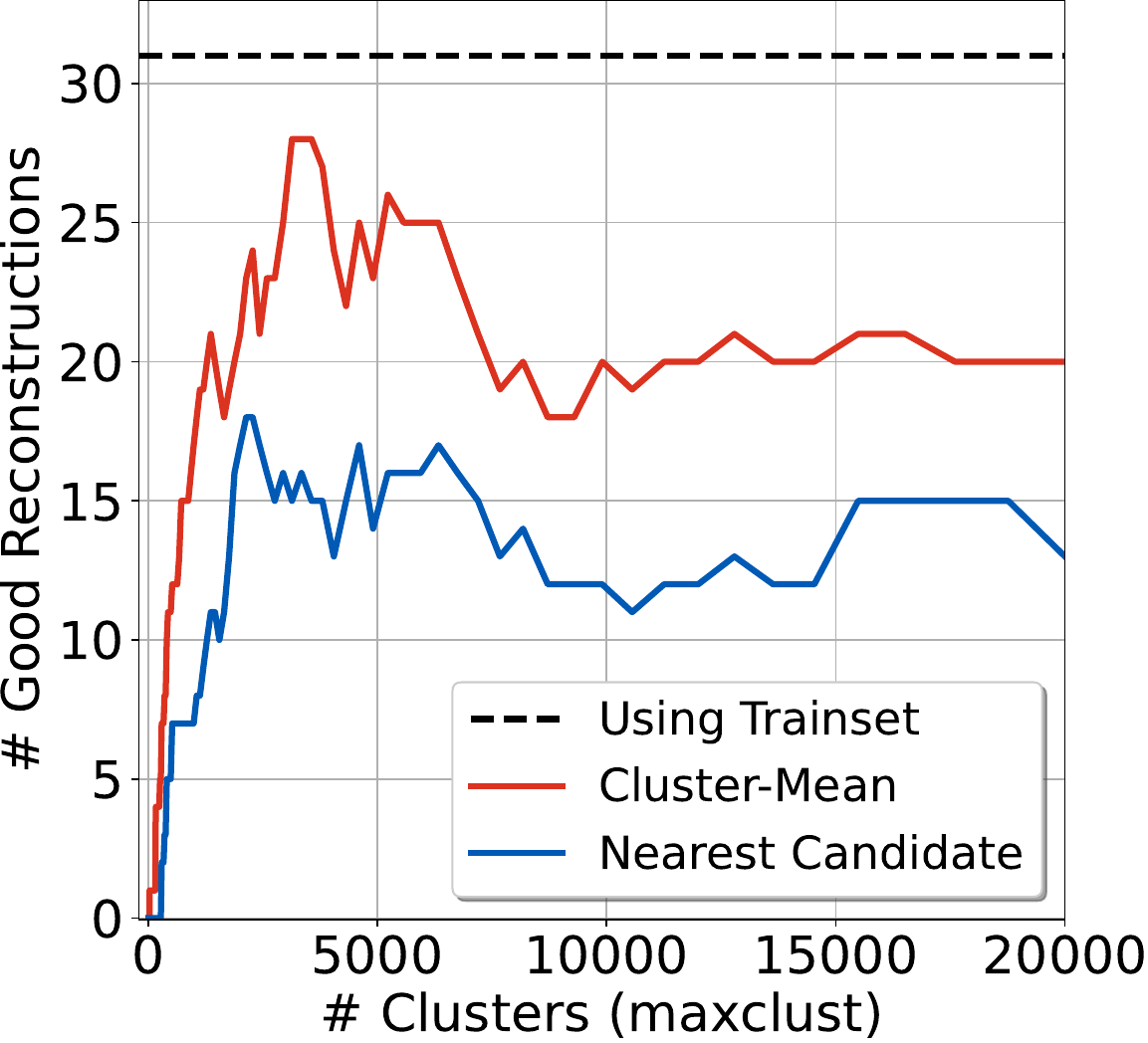}
    \caption{
    Impact of Num. Clusters on Reconstruction Quality (for\\CIFAR10 model with $n$=$500$)
    }
    \label{fig:maxclust}
\end{wrapfigure}

The choice of the number of clusters ({\small \sc maxclust}) significantly affect the results of our clustering-based approach. Since assessing this effect in our current image-embedding setup is computationally prohibitive, we evaluate our approach on $50$k reconstructed candidates from a model trained on $500$ CIFAR-10 images (same as in~\cite{haim2022reconstructing}). For each {\small \sc maxclust}, we select the representatives of the largest $150$ clusters by either averaging all cluster candidates (red) or selecting the nearest candidate to the cluster-mean (blue). We compare each representative to a training image in the same cluster (using SSIM) and count the numbers of good representatives (SSIM>$0.4$), the results are in \cref{fig:maxclust}. 

Notably, beyond a certain small threshold, any {\small \sc maxclust} yields a considerable amount of good reconstructed samples (see also~\cref{appen:clustering}).

\section{Limitations}
\label{sec:limitations}

\newcolumntype{E}{ >{\centering\arraybackslash} m{2.2cm} }
\newcolumntype{F}{ >{\centering\arraybackslash} m{3.2cm} }
\newcommand{\Q}[1]{\multirow{3}{*}{#1}}
\begin{figure}
    \centering
    \hspace*{-2cm}\begin{tabular}{E F F F F}
    \centering
    
    \shortstack{Pretrained\\Model ($\F$)}: & ViT & DINO-ViT & DINOv2-ViT &  CLIP \\
    \toprule
    \raisebox{0pt}{\kern-4em
        \multirow{3}{*}{
            \begin{tabular}[t]{cl}
                Dataset & \\
                \toprule
                \Q{Food101} & $\bs$ \\[.85em]
                & $\F^{-1}(\F(\bs))$ \\[.85em]
                & $\F^{-1}(\hat{\bx})$ \\[1em]
                \Q{iNaturalist} & $\bs$ \\[.85em]
                & $\F^{-1}(\F(\bs))$ \\[.85em]
                & $\F^{-1}(\hat{\bx})$
            \end{tabular}
        }
    } \\
    & \includegraphics[width=.25\textwidth]{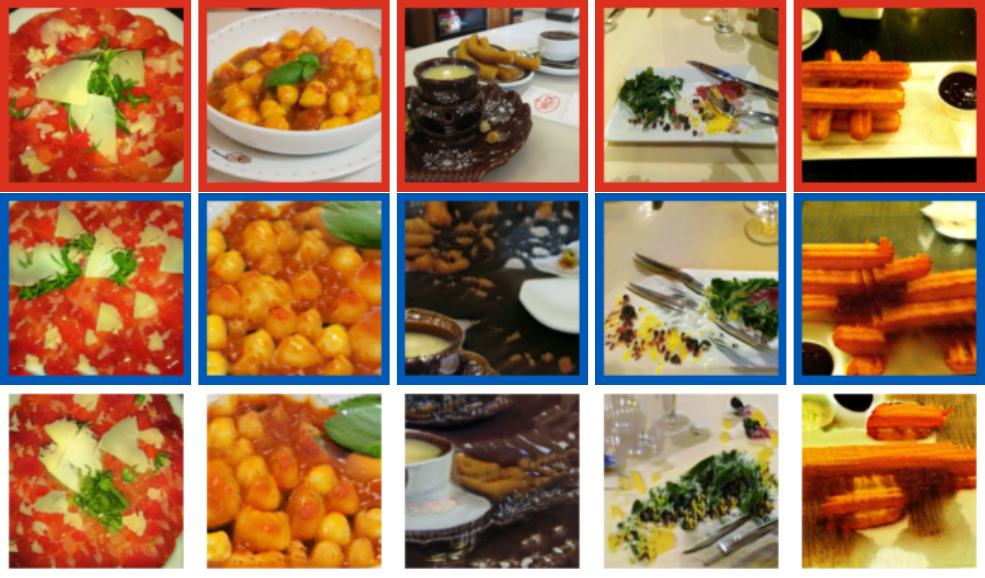}
    & \includegraphics[width=.25\textwidth]{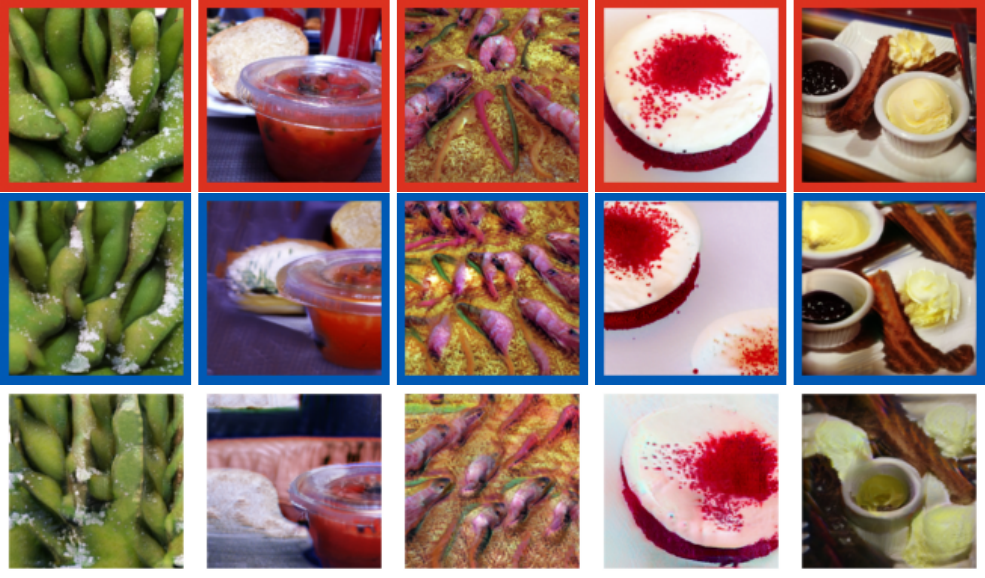} & \includegraphics[width=.25\textwidth]{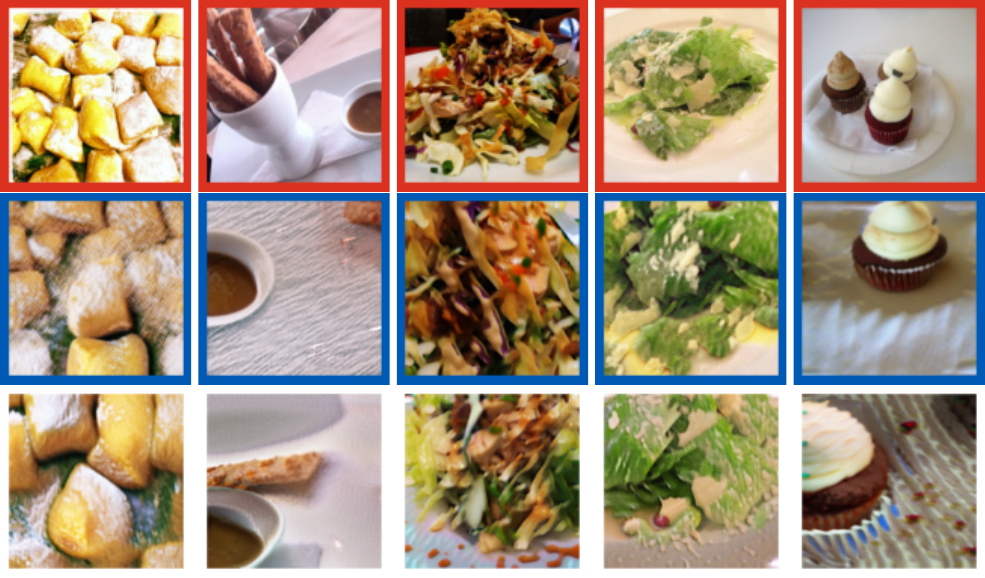} & \includegraphics[width=.25\textwidth]{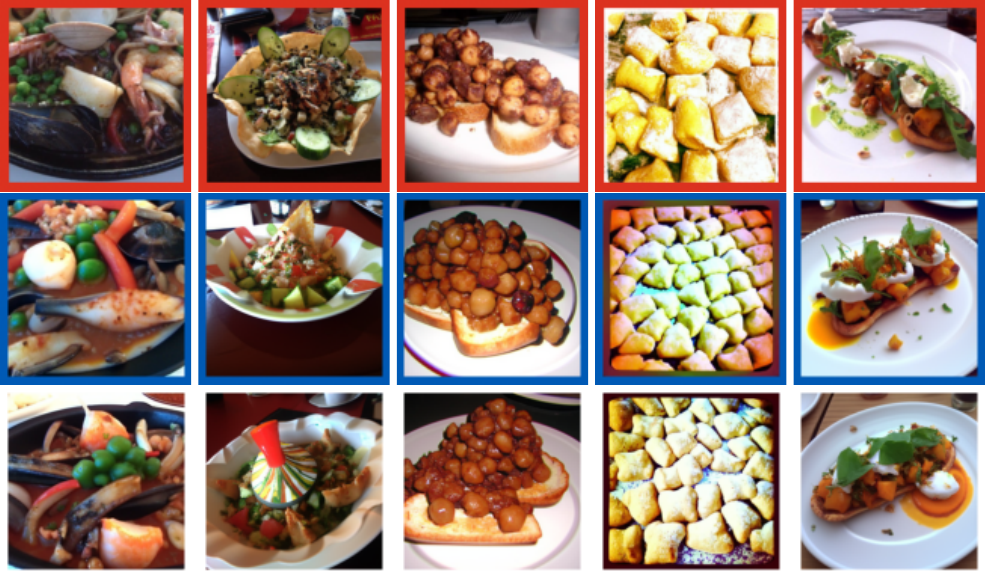} \\

    & \includegraphics[width=.25\textwidth]{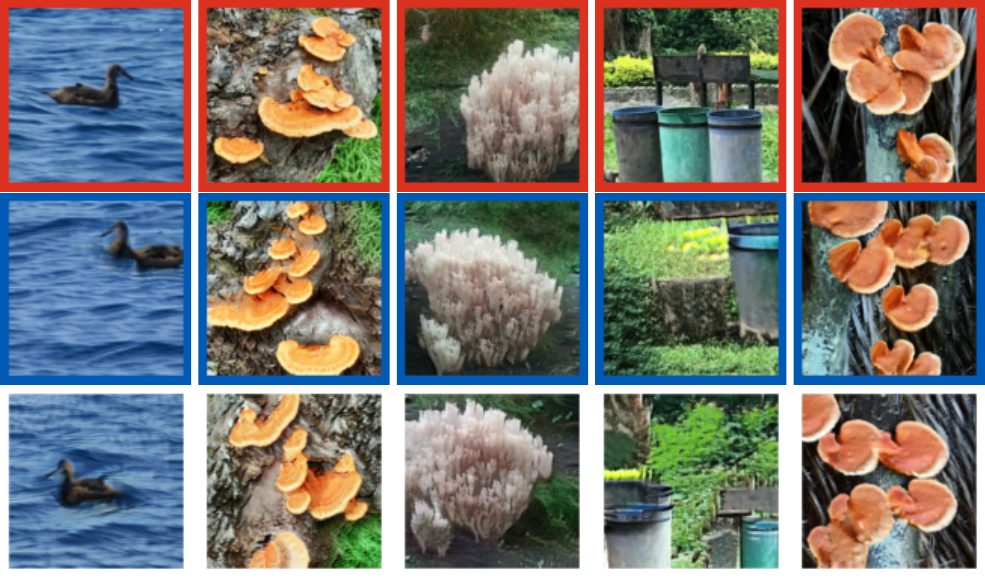} & \includegraphics[width=.25\textwidth]{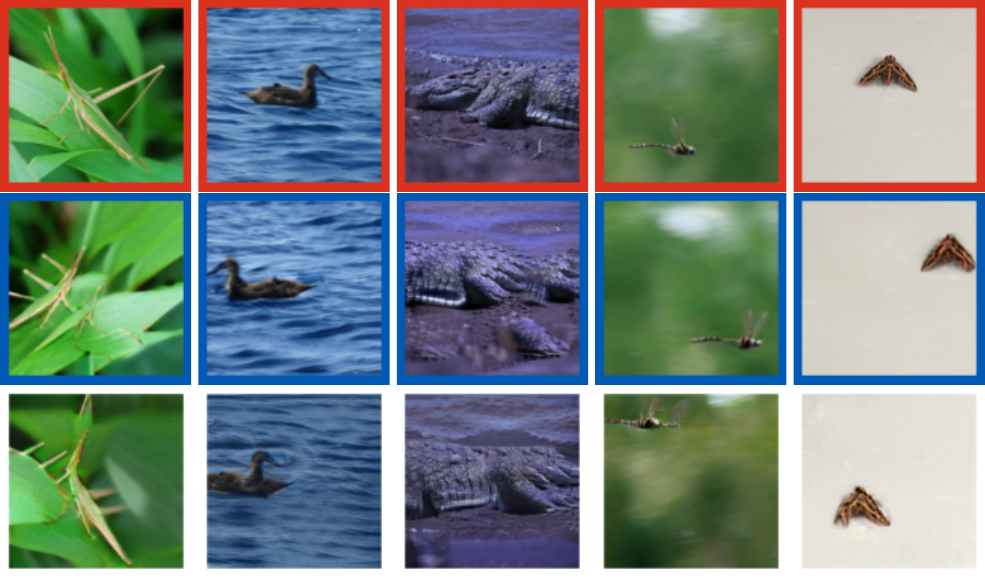} & \includegraphics[width=.25\textwidth]{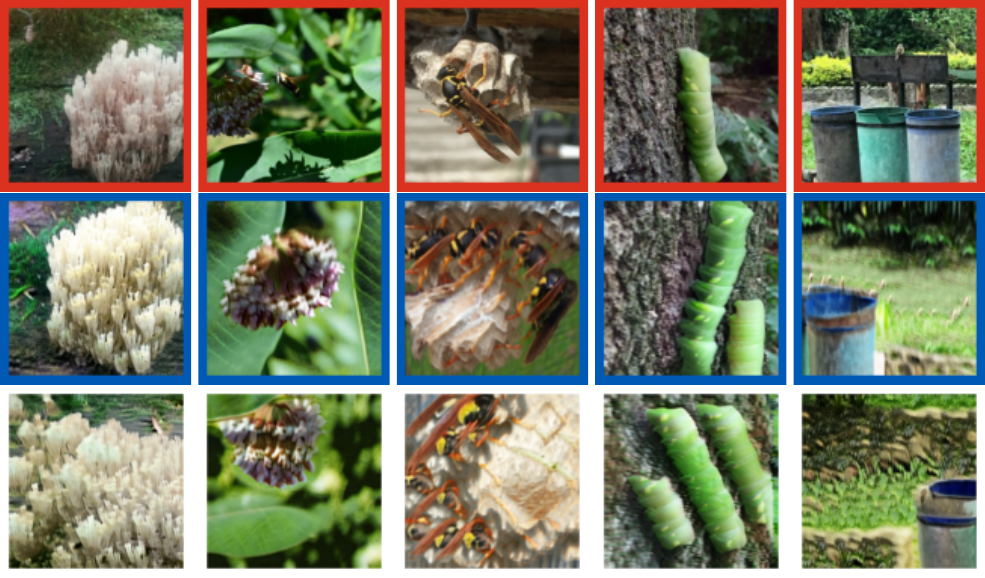} & \includegraphics[width=.25\textwidth]{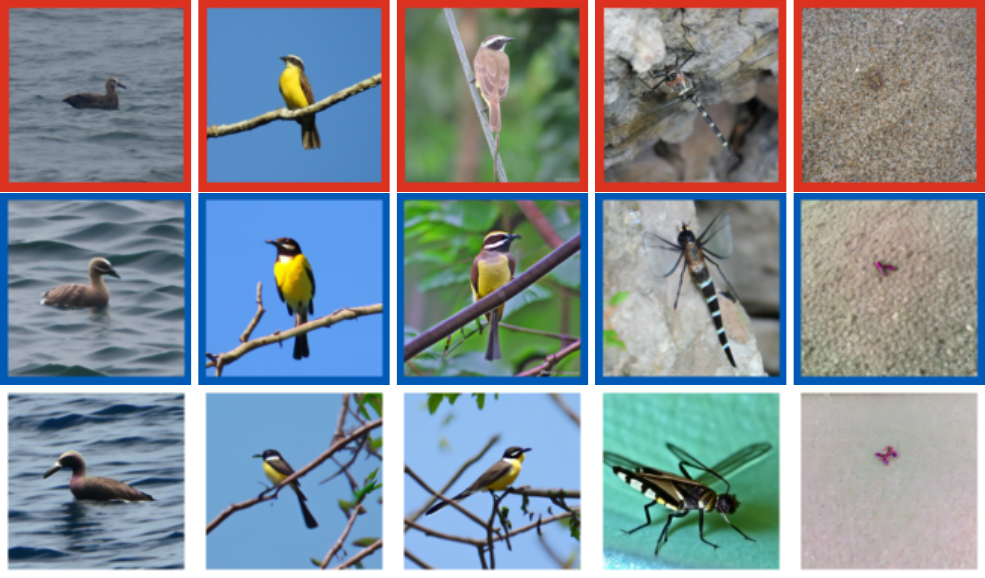} \\
    \end{tabular}
    \caption{Training samples (red), inversion of original embeddings (blue), and inversion of reconstructed embeddings.
    }
    \label{fig:gt_inversion}
\end{figure}

In this work, we made design choices when training the models to align with realistic transfer learning practices. However, some choices led to better reconstruction results than others, revealing limitations of our method. Here we discuss these limitations, their impact on our results, and identify potential future research directions:

\begin{itemize}[leftmargin=12pt]
    \item The quality of reconstructed images relies heavily on the backbone model ($\F$) and the inversion method (\cref{sec:method_inversion}). \cref{fig:gt_inversion} shows the inverted original embeddings $\F^{-1}(\F(\bs))$ (blue), which are the ``best'' we can achieve (independent of our reconstruction method). It also shows how some backbones are easier or harder to invert, as evident in the difference between $\F^{-1}(\F(\bs))$ (blue) and the original image $\bs$ (red), for different $\F$'s. It can also be seen that the inverted reconstructed embeddings $\F^{-1}(\hat{\bx})$ are sometimes more similar to $\F^{-1}(\F(\bs))$ than to $\bs$, which may hint that the challenge lies in the inversion more than in the reconstruction part. Certainly, improving model inversion techniques is likely to enhance the quality of reconstructed samples.
    \item CNN-based backbones $\F$ (e.g., VGG~\citep{simonyan2014very}) proved more challenging for inversion than Transformer-based backbones $\F$. Since Transformers are also being more frequently employed due to their better generalization, we decided to focus our work on them and leave CNN-based backbones for future research.
    \item Linear-Probing (i.e. train a single linear layer $\phi$) is common practice in transfer learning. However, current reconstruction methods, including ours, struggle to perform well on linear models. This may stem from the small number of parameters in linear models (see \cref{appen:samples_vs_width}).
    \item We use weight-decay regularization since it is a fairly common regularization technique. However, the reconstruction method is known to perform much worse on models that are trained without it~\citep{buzaglo2023deconstructing}.
    \item We experimented with an embedding vector that is a concatenation of \cls~ and the average of all other output tokens (of $\F$). This had minor effect on the results, see~\cref{appen:cls_mean_token} for details.
    \item Fine-tuning the entire model $\F$ (together with $\phi$) is resource-intensive and less common compared to training only on fixed embedding vectors. While we followed the latter approach, full fine-tuning can be an interesting future direction.
\end{itemize}

\section{Conclusion}

In this work, we extend previous data reconstruction methods to more realistic transfer learning scenarios. We demonstrate that certain models trained with transfer learning are susceptible to training set reconstruction attack. Given the widespread adoption of transfer learning, our results highlight potential privacy risks. By examining the limitations of our approach, we identify simple mitigation strategies, such as employing smaller or even linear models,  increasing training set size
or training without weight-decay regularization. However, some of these mitigation (removing regularization or using smaller models) may also come at a cost to the generalization of the model. Furthermore, these techniques may not be effective against future advanced reconstruction attacks. 
We aim for our work to inspire the development of new defense methods and emphasize the importance of research on data reconstruction attacks and defenses.

\paragraph*{Acknowledgements.}
We would like to thank Gon Buzaglo for proposing the initial idea of applying data reconstruction to models trained with transfer learning, which laid the foundation for this work. This project is funded by the European Union (ERC grant agreement No 101142115). GV is supported by a research grant from the Center for New Scientists at the Weizmann Institute of Science.

\bibliography{main}

\begin{thebibliography}{44}
\providecommand{\natexlab}[1]{#1}
\providecommand{\url}[1]{\texttt{#1}}
\expandafter\ifx\csname urlstyle\endcsname\relax
  \providecommand{\doi}[1]{doi: #1}\else
  \providecommand{\doi}{doi: \begingroup \urlstyle{rm}\Url}\fi

\bibitem[Biewald(2020)]{wandb}
L.~Biewald.
\newblock Experiment tracking with weights and biases, 2020.
\newblock URL \url{https://www.wandb.com/}.
\newblock Software available from wandb.com.

\bibitem[Bossard et~al.(2014)Bossard, Guillaumin, and Van~Gool]{bossard2014food}
L.~Bossard, M.~Guillaumin, and L.~Van~Gool.
\newblock Food-101--mining discriminative components with random forests.
\newblock In \emph{Computer Vision--ECCV 2014: 13th European Conference, Zurich, Switzerland, September 6-12, 2014, Proceedings, Part VI 13}, pages 446--461. Springer, 2014.

\bibitem[Buzaglo et~al.(2023)Buzaglo, Haim, Yehudai, Vardi, Oz, Nikankin, and Irani]{buzaglo2023deconstructing}
G.~Buzaglo, N.~Haim, G.~Yehudai, G.~Vardi, Y.~Oz, Y.~Nikankin, and M.~Irani.
\newblock Deconstructing data reconstruction: Multiclass, weight decay and general losses.
\newblock In \emph{Advances in Neural Information Processing Systems}, volume~36, pages 51515--51535, 2023.

\bibitem[Carlini et~al.(2019)Carlini, Liu, Erlingsson, Kos, and Song]{carlini2019secret}
N.~Carlini, C.~Liu, {\'U}.~Erlingsson, J.~Kos, and D.~Song.
\newblock The secret sharer: Evaluating and testing unintended memorization in neural networks.
\newblock In \emph{28th USENIX Security Symposium (USENIX Security 19)}, pages 267--284, 2019.

\bibitem[Carlini et~al.(2021)Carlini, Tramer, Wallace, Jagielski, Herbert-Voss, Lee, Roberts, Brown, Song, Erlingsson, et~al.]{carlini2021extracting}
N.~Carlini, F.~Tramer, E.~Wallace, M.~Jagielski, A.~Herbert-Voss, K.~Lee, A.~Roberts, T.~Brown, D.~Song, U.~Erlingsson, et~al.
\newblock Extracting training data from large language models.
\newblock In \emph{30th USENIX Security Symposium (USENIX Security 21)}, pages 2633--2650, 2021.

\bibitem[Carlini et~al.(2023)Carlini, Hayes, Nasr, Jagielski, Sehwag, Tram{\`e}r, Balle, Ippolito, and Wallace]{carlini2023extracting}
N.~Carlini, J.~Hayes, M.~Nasr, M.~Jagielski, V.~Sehwag, F.~Tram{\`e}r, B.~Balle, D.~Ippolito, and E.~Wallace.
\newblock Extracting training data from diffusion models.
\newblock \emph{arXiv preprint arXiv:2301.13188}, 2023.

\bibitem[Caron et~al.(2021)Caron, Touvron, Misra, J{\'e}gou, Mairal, Bojanowski, and Joulin]{caron2021emerging}
M.~Caron, H.~Touvron, I.~Misra, H.~J{\'e}gou, J.~Mairal, P.~Bojanowski, and A.~Joulin.
\newblock Emerging properties in self-supervised vision transformers.
\newblock In \emph{Proceedings of the IEEE/CVF international conference on computer vision}, pages 9650--9660, 2021.

\bibitem[Deng et~al.(2009)Deng, Dong, Socher, Li, Li, and Fei-Fei]{deng2009imagenet}
J.~Deng, W.~Dong, R.~Socher, L.-J. Li, K.~Li, and L.~Fei-Fei.
\newblock Imagenet: A large-scale hierarchical image database.
\newblock In \emph{2009 IEEE conference on computer vision and pattern recognition}, pages 248--255. Ieee, 2009.

\bibitem[Dosovitskiy et~al.(2020)Dosovitskiy, Beyer, Kolesnikov, Weissenborn, Zhai, Unterthiner, Dehghani, Minderer, Heigold, Gelly, et~al.]{dosovitskiy2020image}
A.~Dosovitskiy, L.~Beyer, A.~Kolesnikov, D.~Weissenborn, X.~Zhai, T.~Unterthiner, M.~Dehghani, M.~Minderer, G.~Heigold, S.~Gelly, et~al.
\newblock An image is worth 16x16 words: Transformers for image recognition at scale.
\newblock \emph{arXiv preprint arXiv:2010.11929}, 2020.

\bibitem[Fredrikson et~al.(2015)Fredrikson, Jha, and Ristenpart]{fredrikson2015model}
M.~Fredrikson, S.~Jha, and T.~Ristenpart.
\newblock Model inversion attacks that exploit confidence information and basic countermeasures.
\newblock In \emph{Proceedings of the 22nd ACM SIGSAC conference on computer and communications security}, pages 1322--1333, 2015.

\bibitem[Geiping et~al.(2020)Geiping, Bauermeister, Dr{\"o}ge, and Moeller]{geiping2020inverting}
J.~Geiping, H.~Bauermeister, H.~Dr{\"o}ge, and M.~Moeller.
\newblock Inverting gradients-how easy is it to break privacy in federated learning?
\newblock \emph{Advances in Neural Information Processing Systems}, 33:\penalty0 16937--16947, 2020.

\bibitem[Haim et~al.(2022)Haim, Vardi, Yehudai, Shamir, and Irani]{haim2022reconstructing}
N.~Haim, G.~Vardi, G.~Yehudai, O.~Shamir, and M.~Irani.
\newblock Reconstructing training data from trained neural networks.
\newblock \emph{NeurIPS}, 2022.

\bibitem[He et~al.(2022)He, Chen, Xie, Li, Doll{\'a}r, and Girshick]{he2022masked}
K.~He, X.~Chen, S.~Xie, Y.~Li, P.~Doll{\'a}r, and R.~Girshick.
\newblock Masked autoencoders are scalable vision learners.
\newblock In \emph{Proceedings of the IEEE/CVF conference on computer vision and pattern recognition}, pages 16000--16009, 2022.

\bibitem[He et~al.(2019)He, Zhang, and Lee]{he2019model}
Z.~He, T.~Zhang, and R.~B. Lee.
\newblock Model inversion attacks against collaborative inference.
\newblock In \emph{Proceedings of the 35th Annual Computer Security Applications Conference}, pages 148--162, 2019.

\bibitem[Hitaj et~al.(2017)Hitaj, Ateniese, and Perez-Cruz]{hitaj2017deep}
B.~Hitaj, G.~Ateniese, and F.~Perez-Cruz.
\newblock Deep models under the gan: information leakage from collaborative deep learning.
\newblock In \emph{Proceedings of the 2017 ACM SIGSAC conference on computer and communications security}, pages 603--618, 2017.

\bibitem[Huang et~al.(2021)Huang, Gupta, Song, Li, and Arora]{huang2021evaluating}
Y.~Huang, S.~Gupta, Z.~Song, K.~Li, and S.~Arora.
\newblock Evaluating gradient inversion attacks and defenses in federated learning.
\newblock \emph{Advances in Neural Information Processing Systems}, 34:\penalty0 7232--7241, 2021.

\bibitem[Iman et~al.(2023)Iman, Arabnia, and Rasheed]{iman2023review}
M.~Iman, H.~R. Arabnia, and K.~Rasheed.
\newblock A review of deep transfer learning and recent advancements.
\newblock \emph{Technologies}, 11\penalty0 (2):\penalty0 40, 2023.

\bibitem[Ji and Telgarsky(2020)]{ji2020directional}
Z.~Ji and M.~Telgarsky.
\newblock Directional convergence and alignment in deep learning.
\newblock \emph{Advances in Neural Information Processing Systems}, 33:\penalty0 17176--17186, 2020.

\bibitem[Kim et~al.(2022)Kim, Cosa-Linan, Santhanam, Jannesari, Maros, and Ganslandt]{kim2022transfer}
H.~E. Kim, A.~Cosa-Linan, N.~Santhanam, M.~Jannesari, M.~E. Maros, and T.~Ganslandt.
\newblock Transfer learning for medical image classification: a literature review.
\newblock \emph{BMC medical imaging}, 22\penalty0 (1):\penalty0 69, 2022.

\bibitem[Lee et~al.(2022)Lee, Kim, Choi, Kim, Byeon, Baek, and Kim]{kakaobrain2022karlo-v1-alpha}
D.~Lee, J.~Kim, J.~Choi, J.~Kim, M.~Byeon, W.~Baek, and S.~Kim.
\newblock Karlo-v1.0.alpha on coyo-100m and cc15m.
\newblock \url{https://github.com/kakaobrain/karlo}, 2022.

\bibitem[Loo et~al.(2023)Loo, Hasani, Lechner, and Rus]{loo2023dataset}
N.~Loo, R.~Hasani, M.~Lechner, and D.~Rus.
\newblock Dataset distillation fixes dataset reconstruction attacks.
\newblock \emph{arXiv preprint arXiv:2302.01428}, 2023.

\bibitem[Lyu and Li(2019)]{lyu2019gradient}
K.~Lyu and J.~Li.
\newblock Gradient descent maximizes the margin of homogeneous neural networks.
\newblock \emph{arXiv preprint arXiv:1906.05890}, 2019.

\bibitem[Mahendran and Vedaldi(2016)]{mahendran2016visualizing}
A.~Mahendran and A.~Vedaldi.
\newblock Visualizing deep convolutional neural networks using natural pre-images.
\newblock \emph{International Journal of Computer Vision}, 120:\penalty0 233--255, 2016.

\bibitem[Nasr et~al.(2023)Nasr, Carlini, Hayase, Jagielski, Cooper, Ippolito, Choquette-Choo, Wallace, Tram{\`e}r, and Lee]{nasr2023scalable}
M.~Nasr, N.~Carlini, J.~Hayase, M.~Jagielski, A.~F. Cooper, D.~Ippolito, C.~A. Choquette-Choo, E.~Wallace, F.~Tram{\`e}r, and K.~Lee.
\newblock Scalable extraction of training data from (production) language models.
\newblock \emph{arXiv preprint arXiv:2311.17035}, 2023.

\bibitem[Oquab et~al.(2014)Oquab, Bottou, Laptev, and Sivic]{oquab2014learning}
M.~Oquab, L.~Bottou, I.~Laptev, and J.~Sivic.
\newblock Learning and transferring mid-level image representations using convolutional neural networks.
\newblock In \emph{Proceedings of the IEEE conference on computer vision and pattern recognition}, pages 1717--1724, 2014.

\bibitem[Oquab et~al.(2023)Oquab, Darcet, Moutakanni, Vo, Szafraniec, Khalidov, Fernandez, Haziza, Massa, El-Nouby, et~al.]{oquab2023dinov2}
M.~Oquab, T.~Darcet, T.~Moutakanni, H.~Vo, M.~Szafraniec, V.~Khalidov, P.~Fernandez, D.~Haziza, F.~Massa, A.~El-Nouby, et~al.
\newblock Dinov2: Learning robust visual features without supervision.
\newblock \emph{arXiv preprint arXiv:2304.07193}, 2023.

\bibitem[Paszke et~al.(2019)Paszke, Gross, Massa, Lerer, Bradbury, Chanan, Killeen, Lin, Gimelshein, Antiga, et~al.]{paszke2019pytorch}
A.~Paszke, S.~Gross, F.~Massa, A.~Lerer, J.~Bradbury, G.~Chanan, T.~Killeen, Z.~Lin, N.~Gimelshein, L.~Antiga, et~al.
\newblock Pytorch: An imperative style, high-performance deep learning library.
\newblock \emph{Advances in neural information processing systems}, 32, 2019.

\bibitem[Radford et~al.(2021)Radford, Kim, Hallacy, Ramesh, Goh, Agarwal, Sastry, Askell, Mishkin, Clark, et~al.]{radford2021learning}
A.~Radford, J.~W. Kim, C.~Hallacy, A.~Ramesh, G.~Goh, S.~Agarwal, G.~Sastry, A.~Askell, P.~Mishkin, J.~Clark, et~al.
\newblock Learning transferable visual models from natural language supervision.
\newblock In \emph{International conference on machine learning}, pages 8748--8763. PMLR, 2021.

\bibitem[Ramesh et~al.(2022)Ramesh, Dhariwal, Nichol, Chu, and Chen]{ramesh2022hierarchical}
A.~Ramesh, P.~Dhariwal, A.~Nichol, C.~Chu, and M.~Chen.
\newblock Hierarchical text-conditional image generation with clip latents.
\newblock \emph{arXiv preprint arXiv:2204.06125}, 1\penalty0 (2):\penalty0 3, 2022.

\bibitem[Ronneberger et~al.(2015)Ronneberger, Fischer, and Brox]{ronneberger2015unet}
O.~Ronneberger, P.~Fischer, and T.~Brox.
\newblock U-net: Convolutional networks for biomedical image segmentation.
\newblock In \emph{Medical image computing and computer-assisted intervention--MICCAI 2015: 18th international conference, Munich, Germany, October 5-9, 2015, proceedings, part III 18}, pages 234--241. Springer, 2015.

\bibitem[Simonyan and Zisserman(2014)]{simonyan2014very}
K.~Simonyan and A.~Zisserman.
\newblock Very deep convolutional networks for large-scale image recognition.
\newblock \emph{arXiv preprint arXiv:1409.1556}, 2014.

\bibitem[Somepalli et~al.(2022)Somepalli, Singla, Goldblum, Geiping, and Goldstein]{somepalli2022diffusion}
G.~Somepalli, V.~Singla, M.~Goldblum, J.~Geiping, and T.~Goldstein.
\newblock Diffusion art or digital forgery? investigating data replication in diffusion models.
\newblock \emph{arXiv preprint arXiv:2212.03860}, 2022.

\bibitem[Tan et~al.(2018)Tan, Sun, Kong, Zhang, Yang, and Liu]{tan2018survey}
C.~Tan, F.~Sun, T.~Kong, W.~Zhang, C.~Yang, and C.~Liu.
\newblock A survey on deep transfer learning.
\newblock In \emph{Artificial Neural Networks and Machine Learning--ICANN 2018: 27th International Conference on Artificial Neural Networks, Rhodes, Greece, October 4-7, 2018, Proceedings, Part III 27}, pages 270--279. Springer, 2018.

\bibitem[Tumanyan et~al.(2022)Tumanyan, Bar-Tal, Bagon, and Dekel]{tumanyan2022splicing}
N.~Tumanyan, O.~Bar-Tal, S.~Bagon, and T.~Dekel.
\newblock Splicing vit features for semantic appearance transfer.
\newblock In \emph{Proceedings of the IEEE/CVF Conference on Computer Vision and Pattern Recognition}, pages 10748--10757, 2022.

\bibitem[Ulyanov et~al.(2018)Ulyanov, Vedaldi, and Lempitsky]{ulyanov2018deep}
D.~Ulyanov, A.~Vedaldi, and V.~Lempitsky.
\newblock Deep image prior.
\newblock In \emph{Proceedings of the IEEE conference on computer vision and pattern recognition}, pages 9446--9454, 2018.

\bibitem[Van~Horn et~al.(2018)Van~Horn, Mac~Aodha, Song, Cui, Sun, Shepard, Adam, Perona, and Belongie]{van2018inaturalist}
G.~Van~Horn, O.~Mac~Aodha, Y.~Song, Y.~Cui, C.~Sun, A.~Shepard, H.~Adam, P.~Perona, and S.~Belongie.
\newblock The inaturalist species classification and detection dataset.
\newblock In \emph{Proceedings of the IEEE conference on computer vision and pattern recognition}, pages 8769--8778, 2018.

\bibitem[Wang et~al.(2004)Wang, Bovik, Sheikh, and Simoncelli]{wang2004image}
Z.~Wang, A.~C. Bovik, H.~R. Sheikh, and E.~P. Simoncelli.
\newblock Image quality assessment: from error visibility to structural similarity.
\newblock \emph{IEEE transactions on image processing}, 13\penalty0 (4):\penalty0 600--612, 2004.

\bibitem[Wen et~al.(2022)Wen, Geiping, Fowl, Goldblum, and Goldstein]{wen2022fishing}
Y.~Wen, J.~Geiping, L.~Fowl, M.~Goldblum, and T.~Goldstein.
\newblock Fishing for user data in large-batch federated learning via gradient magnification.
\newblock \emph{arXiv preprint arXiv:2202.00580}, 2022.

\bibitem[Wightman(2019)]{rw2019timm}
R.~Wightman.
\newblock Pytorch image models.
\newblock \url{https://github.com/rwightman/pytorch-image-models}, 2019.

\bibitem[Yang et~al.(2019)Yang, Zhang, Chang, and Liang]{yang2019neural}
Z.~Yang, J.~Zhang, E.-C. Chang, and Z.~Liang.
\newblock Neural network inversion in adversarial setting via background knowledge alignment.
\newblock In \emph{Proceedings of the 2019 ACM SIGSAC Conference on Computer and Communications Security}, pages 225--240, 2019.

\bibitem[Yosinski et~al.(2014)Yosinski, Clune, Bengio, and Lipson]{yosinski2014transferable}
J.~Yosinski, J.~Clune, Y.~Bengio, and H.~Lipson.
\newblock How transferable are features in deep neural networks?
\newblock \emph{Advances in neural information processing systems}, 27, 2014.

\bibitem[Zhang et~al.(2018)Zhang, Isola, Efros, Shechtman, and Wang]{zhang2018perceptual}
R.~Zhang, P.~Isola, A.~A. Efros, E.~Shechtman, and O.~Wang.
\newblock The unreasonable effectiveness of deep features as a perceptual metric.
\newblock In \emph{CVPR}, 2018.

\bibitem[Zhu et~al.(2019)Zhu, Liu, and Han]{zhu2019deep}
L.~Zhu, Z.~Liu, and S.~Han.
\newblock Deep leakage from gradients.
\newblock \emph{Advances in Neural Information Processing Systems}, 32, 2019.

\bibitem[Zhuang et~al.(2020)Zhuang, Qi, Duan, Xi, Zhu, Zhu, Xiong, and He]{zhuang2020comprehensive}
F.~Zhuang, Z.~Qi, K.~Duan, D.~Xi, Y.~Zhu, H.~Zhu, H.~Xiong, and Q.~He.
\newblock A comprehensive survey on transfer learning.
\newblock \emph{Proceedings of the IEEE}, 109\penalty0 (1):\penalty0 43--76, 2020.

\end{thebibliography}
\bibliographystyle{abbrvnat}


\newpage
\appendix

\addcontentsline{toc}{section}{Appendix} 
\part{Appendix} 
\parttoc 

\section{Additional Experiments}

\subsection{Importance of Cosine-Similarity for Inversion (as opposed to MSE)} 
\label{appen:cossim_motivation}

\begin{figure}[h]
    \centering
    \includegraphics[width=0.5\textwidth]{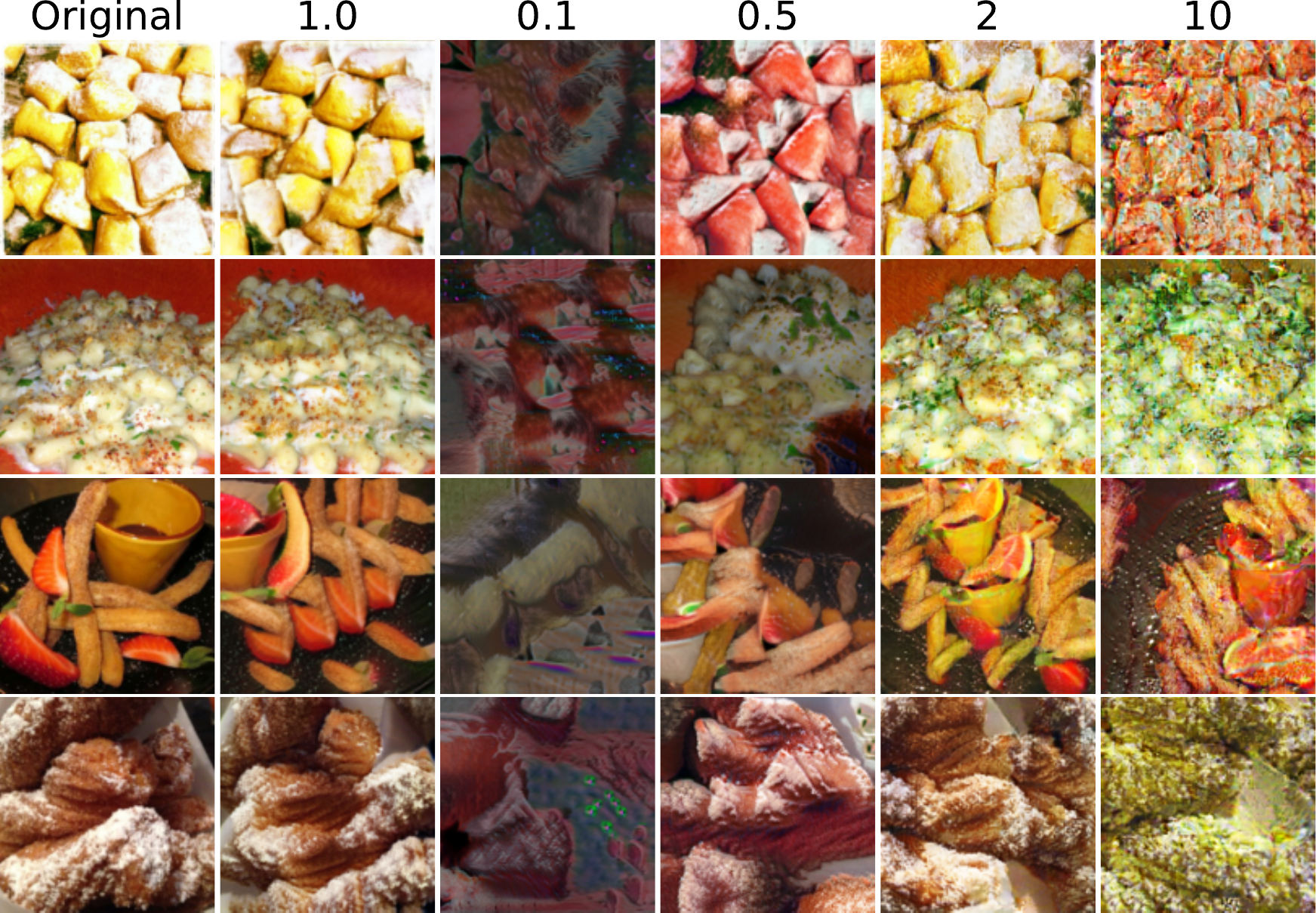}
    \caption{Inverting DINO ($\F^{-1}(a\F(\bs))$) with different scales $a$}
    \label{fig:scales_inversion}
\end{figure}

In~\cref{fig:scales_inversion}, we illustrate the significance of having the correct scale when inverting an embedding (using the inversion described in~\cref{sec:method_inversion}). For several images $\bs$ (left-most column), we display the inversion of their embeddings $\F^{-1}(\F(\bs))$ (second from left column) alongside other inversions of the same vector multiplied by varying scales, namely, $\F^{-1}(a\F(\bs))$ for $a=\left[\frac{1}{10}, \frac{1}{2}, 2, 10\right]$. As clearly evident, inverting the same vector without knowing the "true" scale ($a=1.0$) would result in very different results, sometimes making them hard to recognize. 

The original paper~\cite{tumanyan2022splicing} uses MSE in its inversion scheme. However, the output candidates from the reconstruction method (described in~\cref{sec:method_reconstruction}) can have significantly different norms than their corresponding original training embedding.

\begin{figure}[tbhp]
    \centering
    \begin{tabular}{cc}
         \includegraphics[width=0.48\textwidth]{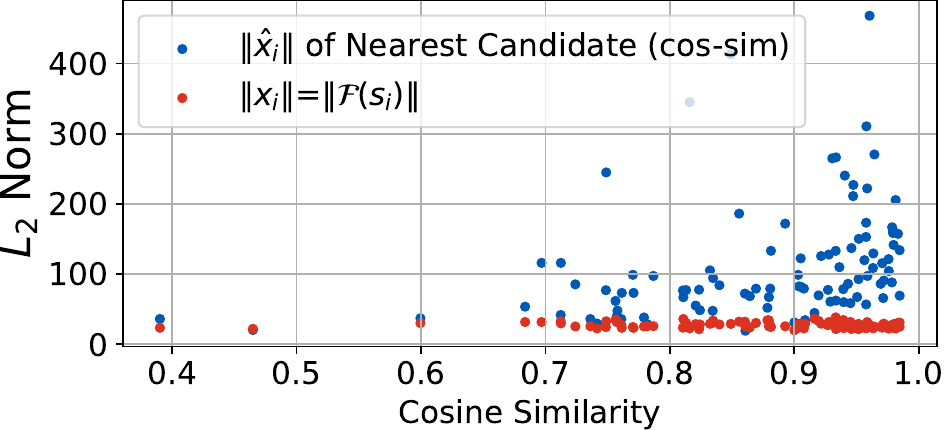}
         &
         \includegraphics[width=0.48\textwidth]{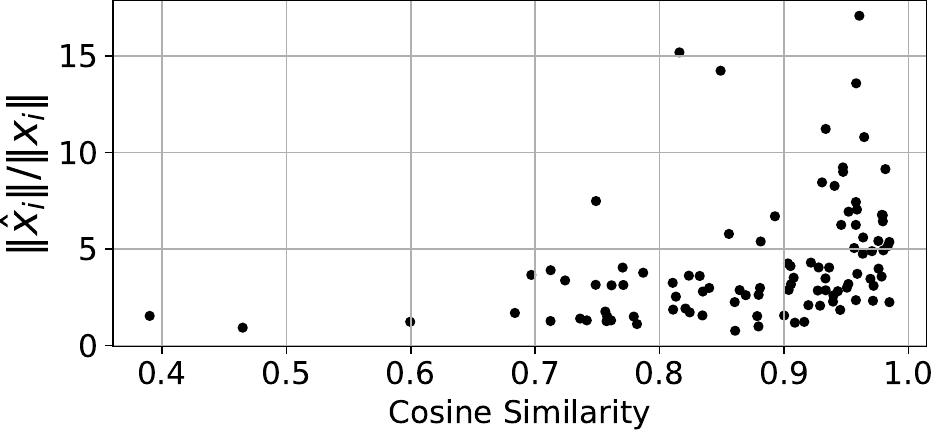} \\
         (a) & (b) \\
    \end{tabular}
    \caption{Comparing (a) the norms of $\F({\bs})$ (red) and its NN $\hat{\bx}$ (blue), and (b) their ratios}
    \label{fig:scale_mse_cossim}
\end{figure}

To conduct a comparison, we employ a binary model trained on DINO embeddings of images from Food101, reconstructing candidates $\hat{\bx}$ from this model. In~\cref{fig:scale_mse_cossim}a, for each training image $\bs$ we compare the norm of its DINO embedding $\Vert \F(\bs) \Vert$ (red), to the norm of its nearest neighbour embedding $\Vert \hat{\bx} \Vert$ (blue), where $\hat{\bx} = \underset{\bx}{\text{argmin}}~d_{cosine}(x, \F(\bs))$ (and the value of $d_{cosine}$ is the x-axis). In~\cref{fig:scale_mse_cossim}b we show the ratio between the two, highlighting that candidates can have very different norm compared to their corresponding training image. This variation in norms is a result of the reconstruction scheme that we use~(\cref{sec:method_reconstruction}), whose nature we don't fully understand yet. However, using cosine-similarity loss in our inversion scheme eliminates this issue.

\subsection{Cosine-Similarity as a proxy for Good Reconstructions}
\label{appen:cossim_as_proxy}

Determining whether a candidate is a reconstruction of an original sample is a difficult challenge. It is highly unlikely that the two will be exactly the same, which is why selecting an appropriate similarity measure is important. Unfortunately, there is no "best" metric for comparing two images, which is a known open problem in computer vision (see e.g., \cite{zhang2018perceptual}).

The task becomes even more complex when dealing with reconstructed embedding vectors, as in our work. Given our computational constraints, we must choose wisely which embeddings to invert, adding another layer of complexity to the comparison process. Throughout our work, we frequently employ cosine similarity as a metric for evaluating embedding similarities. However, whether this metric accurately reflects visual quality is unclear. We set out to explore this question empirically.

Previous work on data reconstruction~\citep{haim2022reconstructing,buzaglo2023deconstructing} directly reconstruct training images, allowing us to a directly compare between cosine similarity and image similarity measures. Both works established SSIM~\citep{wang2004image} as a good visual metric for CIFAR10 images (see Appendix~A.2 in~\cite{buzaglo2023deconstructing}), and defined SSIM$>0.4$ as a good threshold for declaring two images as sufficiently similar. In fact these works also use cosine similarity to find nearest neighbors between candidates and training images when normalized to $[-1,1]$, and only after shifting the images to [0,1] they use SSIM. Which means that they also implicitly assume that cosine-similarity is a good proxy for visual similarity.

\begin{figure}[h]
    \centering
    \includegraphics[width=0.85\textwidth]{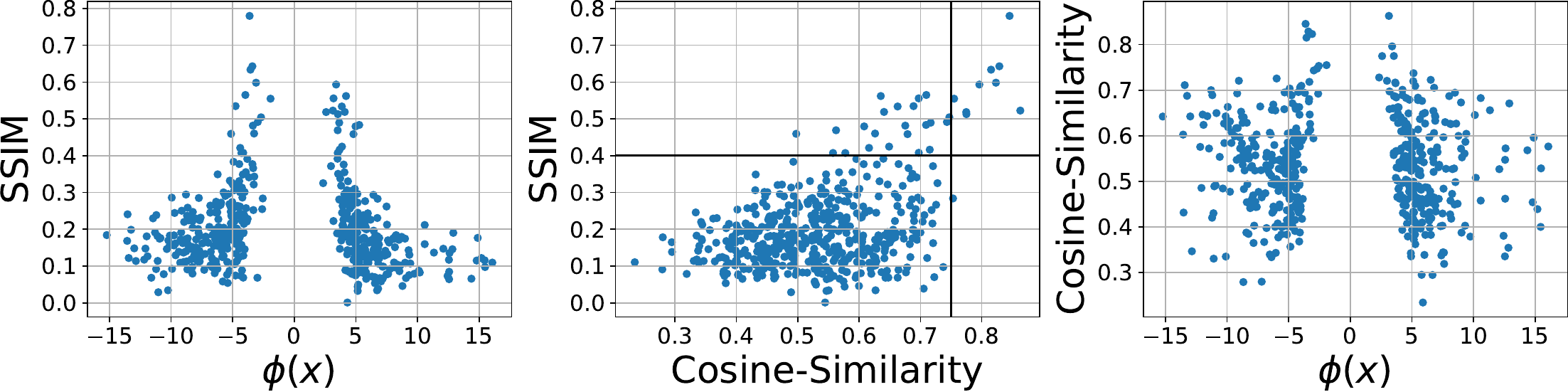}
    \caption{Comparing Cos-Sim to SSIM of training data (model trained on CIFAR10)}
    \label{fig:cifar_ssim_vs_cossim}
\end{figure}

In~\cref{fig:cifar_ssim_vs_cossim}, we quantitatively evaluate this assumption using reconstructed images from a CIFAR10-trained model (as in~\cite{haim2022reconstructing}). The left panel is for simply reproducing the results of~\cite{haim2022reconstructing}. By looking at both middle and right panel, we see that CosSim=$0.75$ is a good cut-off for determining "good" reconstruction, since from this point there is a good correlation between the two metrics. This is also the reason that we use this threshold for determining good reconstruction in other experiments in the paper.

By further observing the middle panel: if SSIM$>0.4$ (horizontal black line) is considered a criterion for good image reconstruction, then cosine similarity (CosSim$>0.75$, vertical black line) may overlook some potentially high-quality reconstructions, indicating room for further improvement in our approach.

\subsection{Does Training Data Reconstructability Require Overtraining?}
\label{appen:iterations}

\begin{figure}[htbp]
    \includegraphics[width=\textwidth]{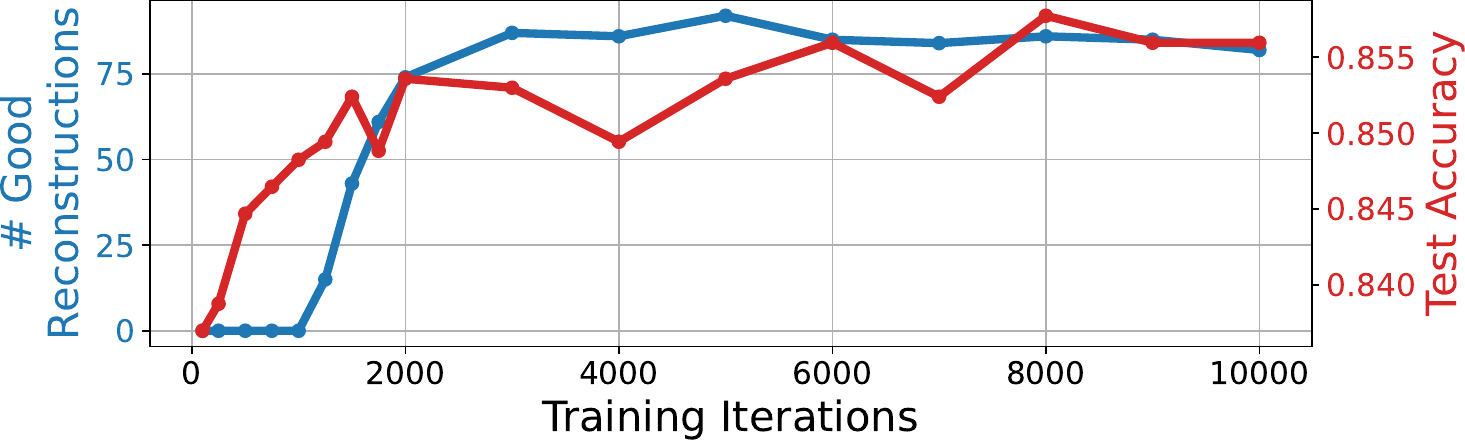}
    \caption{Does Training Data Reconstructability Require Overtraining? -- Seems Not.}
    \label{fig:iterations_test_vs_good}
\end{figure}

We set to explore how "reconstructability" (i.e., how many good samples we can reconstruct from) depends on the number of training iterations. We note from empirical observations that reconstructability certainly improves with longer training, which should not be surprising because according to theory, the model converges more to the KKT solution. 

But the key question is - does the model have to be "overtrained" before becoming reconstructable, or not? To define "overtrained", we observe how the generalization accuracy increases. Obviously, the longer we train, the better the model will be reconstructable. But is it reconstructable before the generalization accuracy saturates? (Or do we have to keep training long after that?)

In~\cref{fig:iterations_test_vs_good} we show the test accuracy per training iteration (red) for a model trained on DINO embeddings from the Food101 dataset. We also show reconstruction quality (blue) by counting the number of training samples whose cosine similarity to its nearest neighbor candidate was above $0.75$. As can be seen, reconstructability increases after about 1000 iterations and starts saturating at about 2000 iterations, where the test accuracy (even though quite high in the beginning), keeps increasing by more than 1.5\% until 10k iterations.

The implication is that reconstructability is achieved in a reasonable time (measured by the time taken to achieve good generalization accuracy). This observation is important to assert the realism of our method as a viable privacy threat to models trained in a similar fashion.

\subsection{Impact of Model Size and Training Set Size on Reconstructability}
\label{appen:samples_vs_width}

\begin{figure}[htbp]
    \centering
    \includegraphics[width=.7\textwidth]{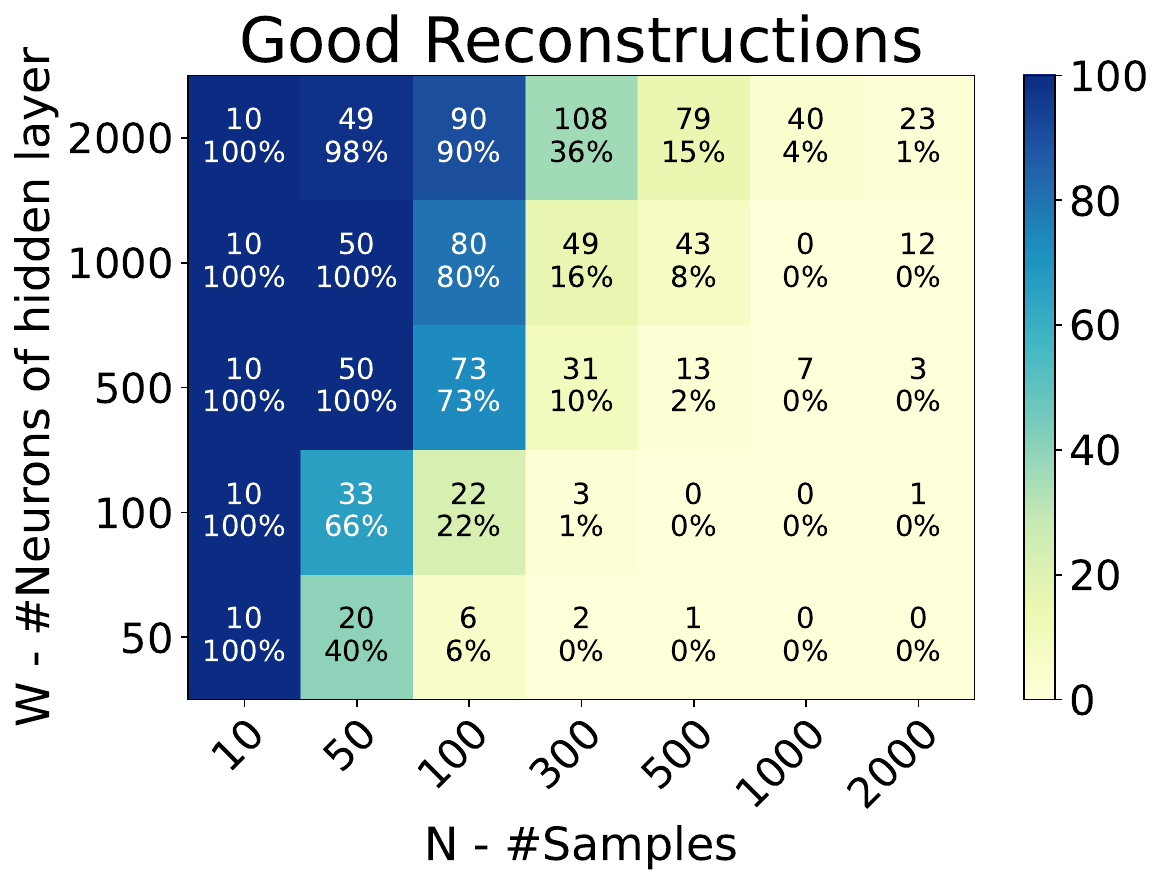}
    \caption{Effect of model size and dataset size on reconstructability.}
    \label{fig:samples_vs_width}
\end{figure}

Previous works~\cite{buzaglo2023deconstructing} observed that the quality of reconstruction results is influenced by the size of the model (i.e., number of parameters) and the size of the training set. We conduct similar analysis for our models. 

This relationship can be intuitively understood by considering ~\cref{eq:reconstruction_loss} as a system of equations to be inverted, where the number of equations corresponds to the number of parameters in the model, $\theta \in \reals^p$, and the unknowns are the coefficients $\lambda_i \in \reals$ and the reconstructed embeddings ${\bx}_i \in \reals^d$ for each training sample $i \in \{1,...,n\}$. The ratio $\frac{p}{n(d+1)}$ represents the number of model parameters relative to the total number of unknowns. As this ratio increases, i.e., when the model has more parameters compared to the number of unknowns, we hypothesize that the system of equations becomes more well-determined, leading to higher reconstructability. 

This hypothesis is supported by the empirical results presented in~\cref{fig:samples_vs_width}, where we train $2$-layer MLPs with architecture $D$-$W$-$1$ on $N$ training samples from binary Food101. Each cell reports the number of good reconstructions (cosine similarity between training embedding and its nearest neighbor candidate $> 0.75$), both in absolute terms and as a percentage relative to N. As shown, when the model has more parameters relative to the number of training examples (further left and higher up in the table), our method can extract more reconstructions from the model.

This figure also show that our method can be extended to larger datasets, up to $N$=$2000$ (and probably beyond).

\subsection{Effect of Using \cls+MEAN vs \cls~ as Feature Vector}
\label{appen:cls_mean_token}

\begin{figure}[tbhp]
    \centering
    \includegraphics[width=\textwidth]{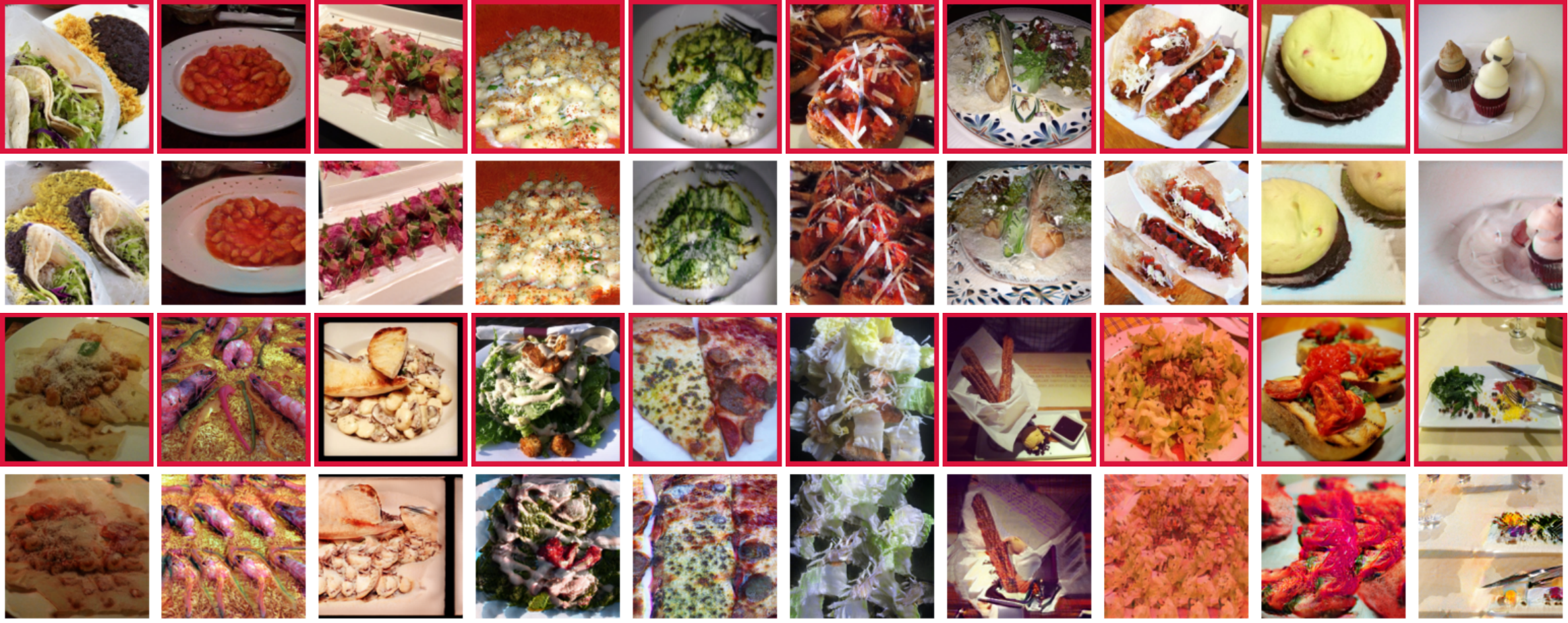}
    \caption{Reconstruction from a DINO model trained on \cls+MEAN embedding vector (original training image in red)}
    \label{fig:cls_mean_dino_food}
\end{figure}

In our work we use the \cls~ token as the feature vector for a given image. However, there may be other ways to use the outputs of transformer-based foundation models as feature vectors. As suggested in \cite{caron2021emerging} (linear probing section), one might use a concatenation of the \cls~ token and the mean of the rest of the other output tokens (\cls+MEAN). In \cref{fig:cls_mean_dino_food} we show reconstructed results for a model that was trained using such \cls+MEAN feature vector (using DINO on Food101). As seen, the extra information in the feature vector does not seem to have a significant effect on the total results of the reconstruction (as opposed to a possible assumption that extra information would result in higher reconstructability). While this is by no means an exhaustive evaluation of this design choice (using \cls+MEAN vs. just \cls), it does look like this may not change the results of the reconstruction too much.

\subsection{Model Inversion for CLIP vs UnCLIP Decoder}
\label{appen:clip_dip_inversion}

\begin{figure}[thbp]
    \centering
    \begin{tabular}{cc}
         \includegraphics[width=0.45\textwidth]{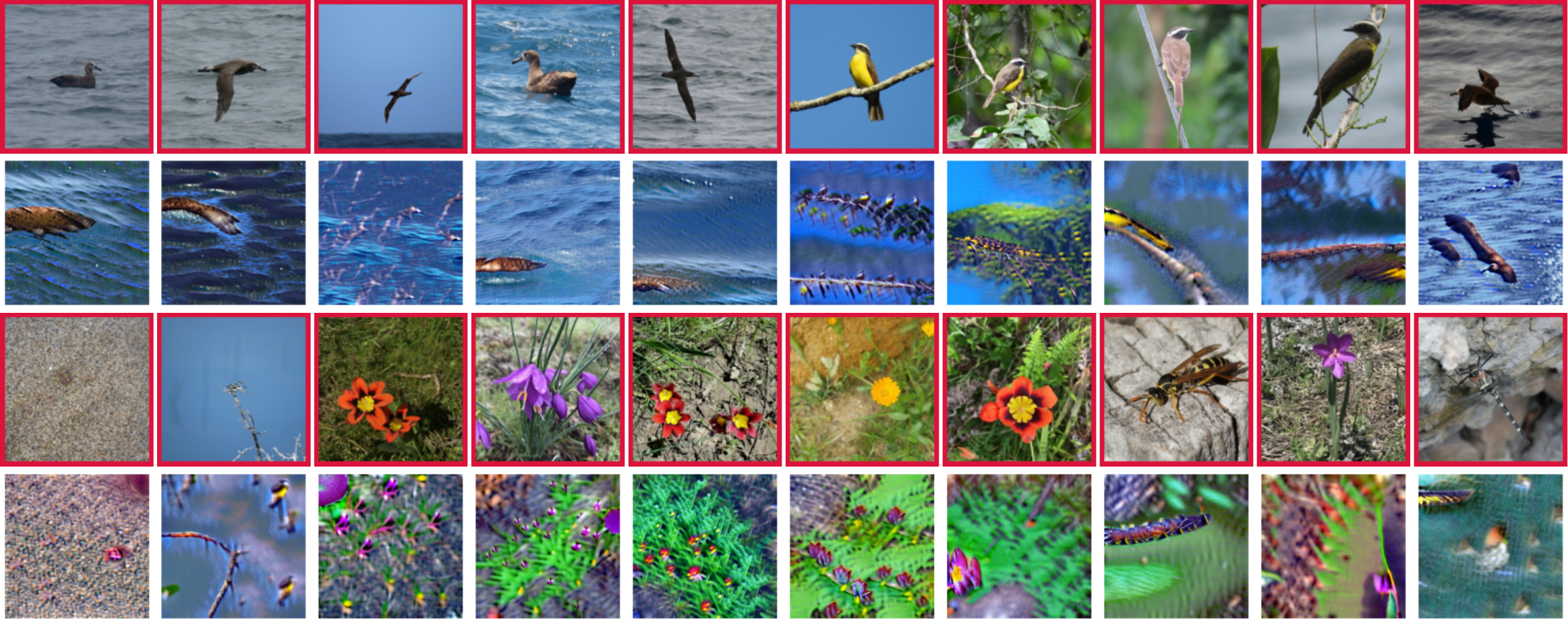}&
         \includegraphics[width=0.45\textwidth]{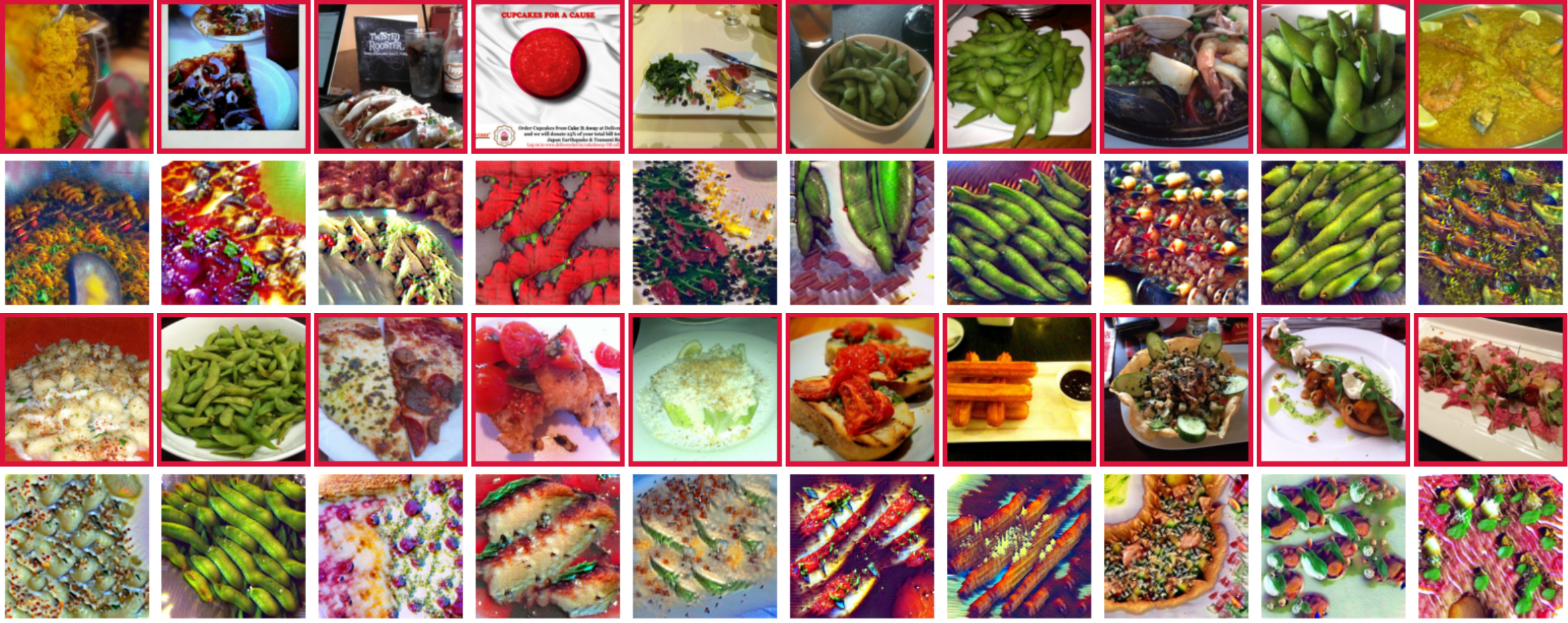} \\
         (a) iNaturalist & (b) Food101
    \end{tabular}
    \caption{Model-Inversion reconstructions from a model trained on CLIP embeddings}
    \label{fig:clip_dip}
\end{figure}

As described in~\cref{sec:method_inversion}, for inverting CLIP embeddings, we use an UnCLIP decoder instead of the model inversion approach used for other backbone models (ViT/DINO/DINOv2). The main reason behind this choice is that the same inversion method did not seem to provide satisfactory results for CLIP. In~\cref{fig:clip_dip}, we show output images of inverted embeddings using the approach from \cite{tumanyan2022splicing} (with the modifications described in our paper). The results do not produce comparable quality to using the UnCLIP decoder.

\subsection{Further Insights on Clustering-based Reconstruction (\cref{sec:clustering})}
\label{appen:clustering}

\begin{figure}[tbhp]
    \centering
    \includegraphics[width=\textwidth]{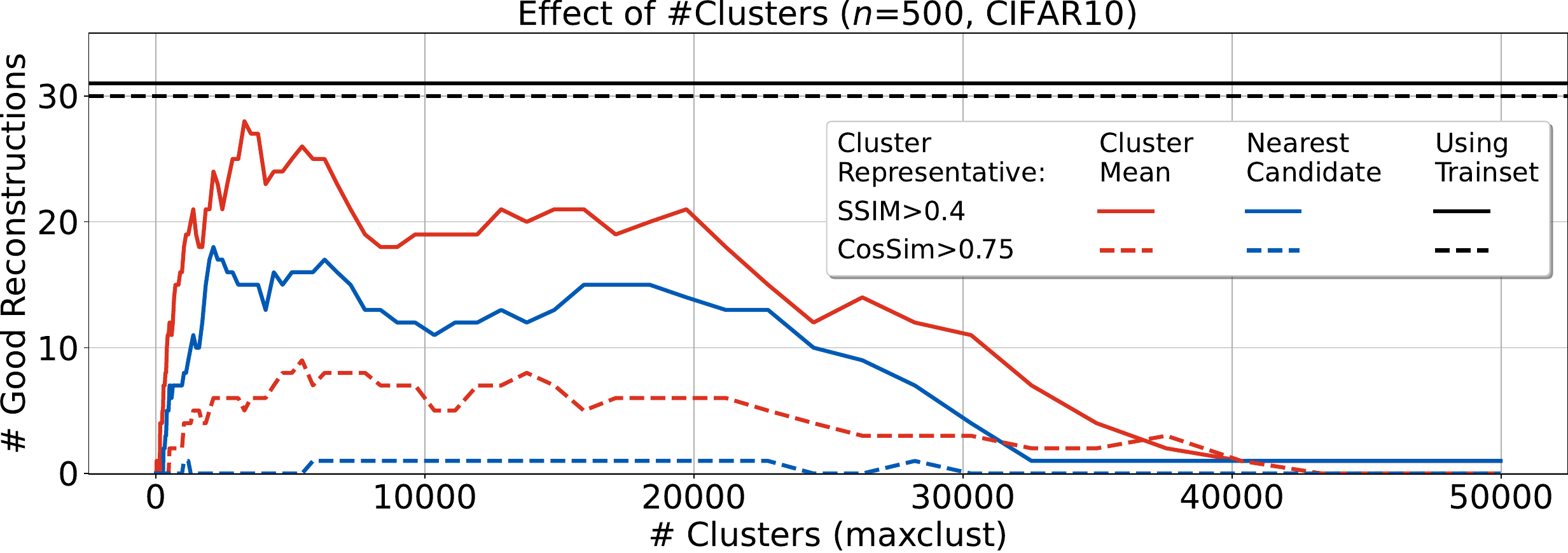}
    \caption{Extended Results for the figure in~\cref{sec:clustering}}
    \label{fig:clustering_full}
\end{figure}

In~\cref{fig:clustering_full}, we show extended results of the inset Figure in~\cref{sec:clustering}, displaying the same graph up to larger {\small \sc maxclust} values (red and blue solid lines), together with similar results that count the number of ``good'' reconstructions with CosSim $>0.75$ (dashed blue and red lines).

The reason for the decrease in the number of good reconstructions as the number of clusters increases, is that we only consider the largest $150$ clusters (per partition of the candidates, as determined by {\small \sc maxclust}). Consequently, when there are too many clusters, the probability that the largest ones correspond to a cluster of a training sample decreases (for $50$k clusters, this becomes totally random). Note that the largest number of clusters in the graph is slightly smaller than 50k, and there exist several clusters with 2-3 candidates.

Another insight from this graph is that averaging several candidates together results in better candidates, an observation also made by \cite{haim2022reconstructing}. In our work, we don't use such candidate averaging (except for the clustering experiments), but this may lead to improved results. We leave this for future research.

We note that since the similarity measure between candidates is cosine similarity, this implicitly applies a spherical topography for comparing candidates. Therefore, it is not straightforward to compute the mean of several candidates. In our work, we use the simple arithmetic mean, which empirically seems to work well. We considered computing the Fréchet mean, i.e., the mean of the candidates that lies on the sphere, but could not find a working implementation for this. This may also be an interesting direction for future research.

\begin{figure}[tbhp]
    \centering
    \includegraphics[width=\textwidth]{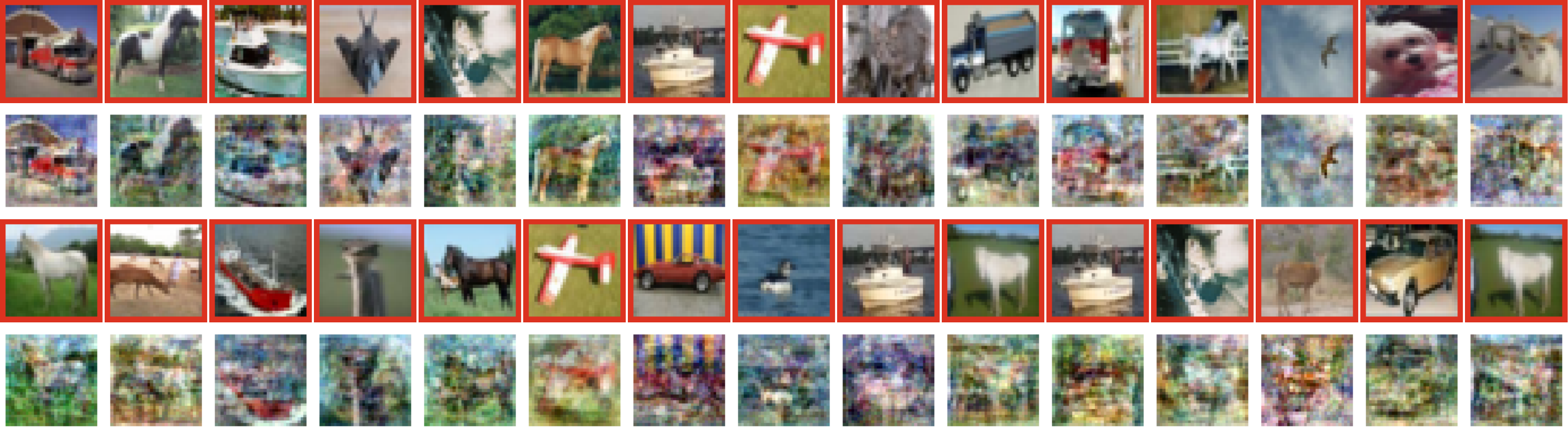}
    \caption{Reconstructed candidates of CIFAR10 model, obtained with our clustering-based approach for the "peak" value in~\cref{fig:clustering_full} ($3294$)}
    \label{fig:clustering_full_images}
\end{figure}

For completeness, we show how the reconstructed samples look for the choice of the "peak" SSIM from~\cref{fig:clustering_full}, which occurs at 3294 clusters. These are shown in~\cref{fig:clustering_full_images}.

\subsection{Comparison to Activation Maximization}
\label{appen:activation_maximization}

\begin{figure}[tbhp]
    \centering
    \includegraphics[width=\textwidth]{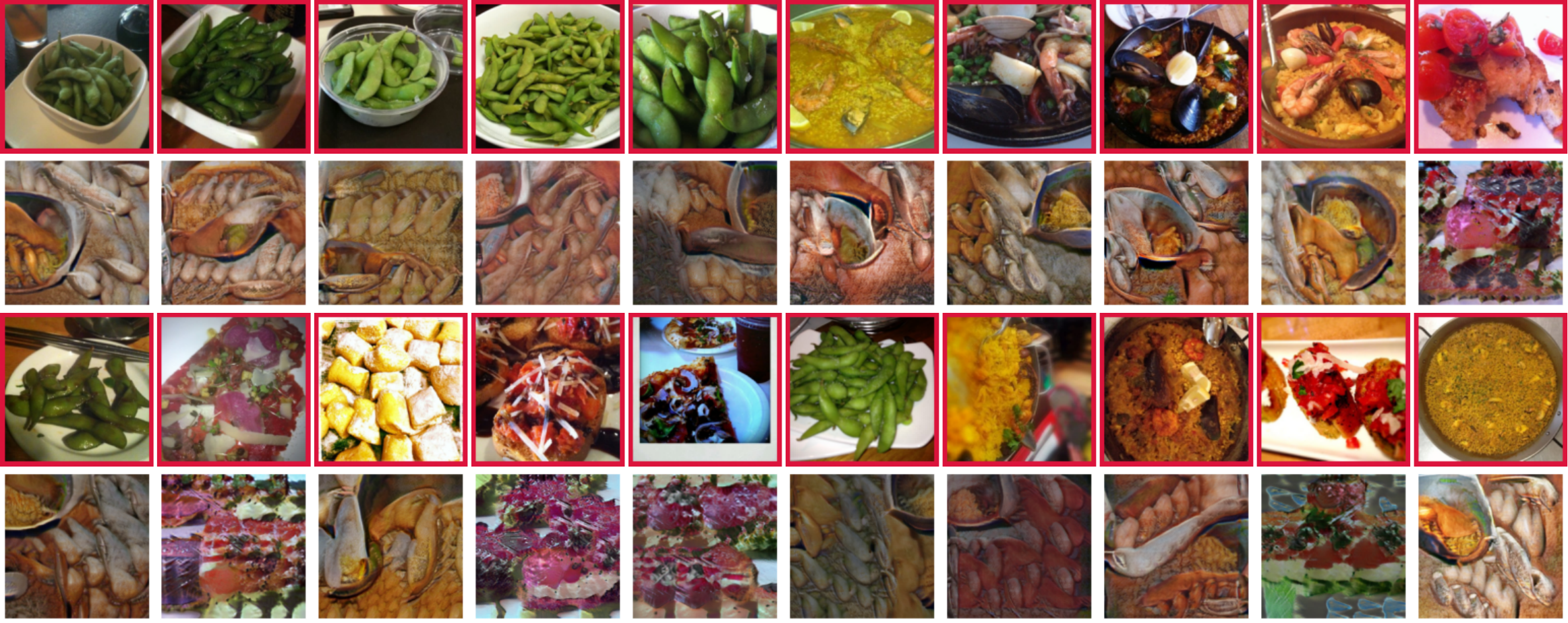}
    \caption{Reconstructions using activation maximization on the input to $\phi$}
    \label{fig:activation_food}
\end{figure}

\begin{figure}[tbhp]
    \centering
    \includegraphics[width=\textwidth]{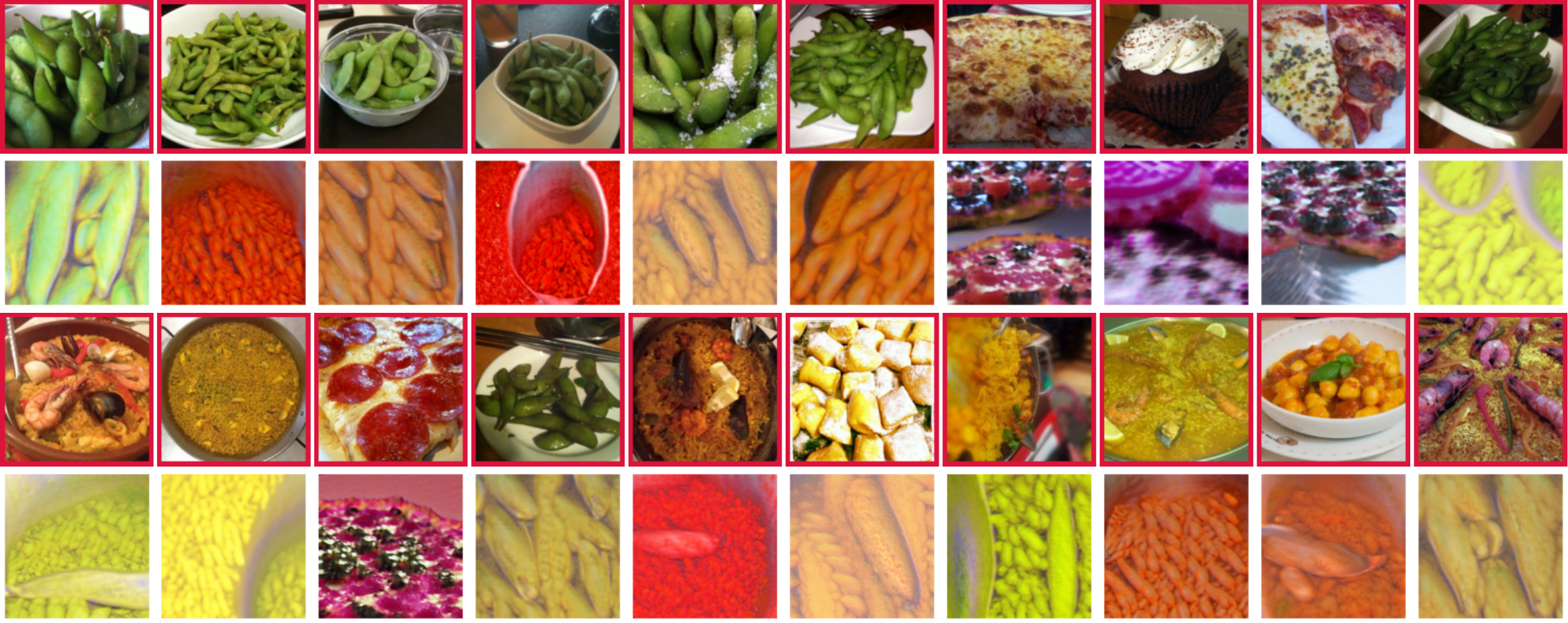}
    \caption{Reconstructions using activation maximization on the input to $\F$}
    \label{fig:activation_base_food}
\end{figure}

We compare our reconstruction results to a popular baseline for reconstructing data from trained model. It is called ``model inversion'' in the context of privacy\citep{fredrikson2015model} or ``activation maximization'' in the context of visualization techniques~\citep{mahendran2016visualizing} (we prefer the term activation maximization as it is more accurate). We are searching for inputs to the model that achieve high activations for the model's outputs that correspond to each class. We consider two options in our case:

 The first, by performing activation maximization on the inputs to $\phi$:
\[
\underset{\bx}{\text{argmin}}\ \mathcal{L} \left(\Phi(\bx),\ y\right)
\]

This results in multiple candidates $\{\bx\}$ that minimize the loss function (binary cross-entropy) w.r.t to the classes $y \in \{-1,1\}$. We then search for candidates that are nearest neighbours of original training embeddings, and invert them to images by computing $\F^{-1}(\bx)$ (this is the same pipeline as we use for the reconstructed candidates of our approach). The results of this approach can be seen in~\cref{fig:activation_food}.

The second approach, is to optimize over the inputs to $\F$ (instead of the inputs to $\phi$) in the same manner that is described in~\cref{appen:imp_inversion}:
\[
\bx=g_{\nu}(z)~~~ s.t.~~ \nu = \underset{\nu}{\text{argmin}}\ \mathcal{L} \left( \phi\left( \F \left( g_{\nu}(z) \right) \right), y\right)
\]

Where $g$ is the U-Net model with parameters $\nu$. This is equivalent to performing the inversion method as described in~\cref{appen:imp_inversion}, but feeding the output of $\F$ into the trained $\phi$ and then into the loss $\mathcal{L}$, with a given $y \in \{-1,1\}$ (instead of comparing the output of $\F$ to a given embedding vector). Note that as described in~\cref{appen:imp_inversion}, the only optimization variables are the parameters of the Deep-Image Prior $g$ (denoted as $\nu$ in the equation above). The results of this approach are shown in~\cref{fig:activation_base_food}.

As evident from the results, while activation maximization techniques manage to reconstruct some interesting outputs, that are somewhat semantically related to the training classes, the results of both methods are inferior to the results of our proposed approach.

\subsection{GT Inversion} 
\label{appen:gt_inversion}

Here are the full results (on all reconstructed candidates) of the results shown in~\cref{fig:gt_inversion}.

\begin{figure}
    \centering
    \includegraphics[width=\textwidth]{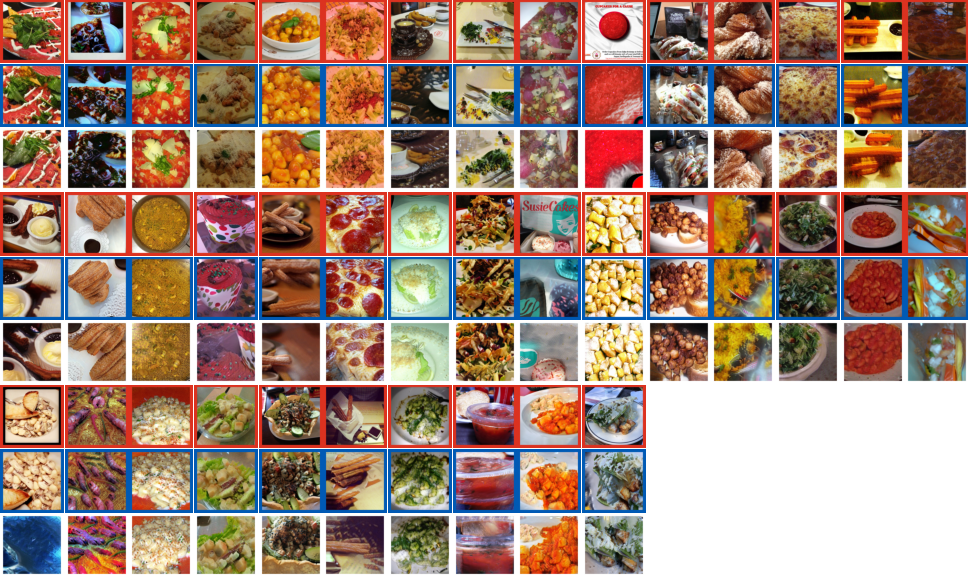}
    \caption{ViT on Food101: Training Image (red), Inversion of Original Embedding (blue) and Inversion of Reconstructed Embedding.}
    \label{fig:gt_inversion_vit_food}
\end{figure}

\begin{figure}
    \centering
    \includegraphics[width=\textwidth]{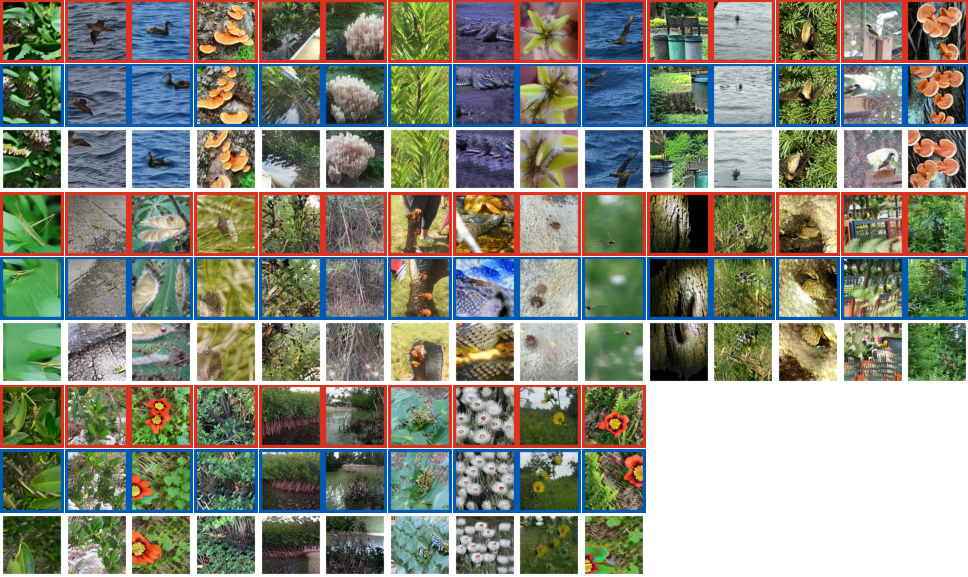}
    \caption{ViT on iNaturalist: Training Image (red), Inversion of Original Embedding (blue) and Inversion of Reconstructed Embedding.}
    \label{fig:gt_inversion_vit_inat}
\end{figure}

\begin{figure}
    \centering
    \includegraphics[width=\textwidth]{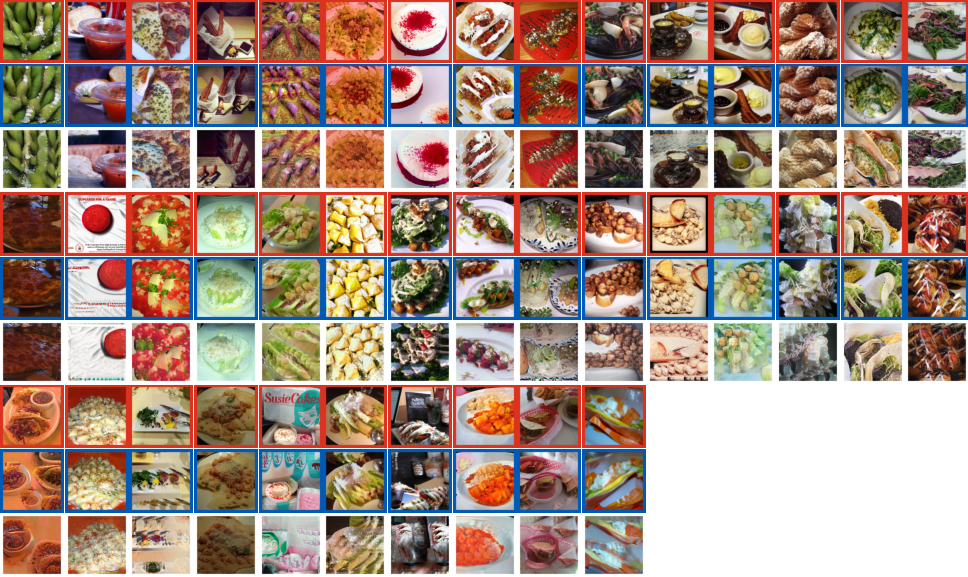}
    \caption{DINO on Food101: Training Image (red), Inversion of Original Embedding (blue) and Inversion of Reconstructed Embedding.}
    \label{fig:gt_inversion_dino_food}
\end{figure}

\begin{figure}
    \centering
    \includegraphics[width=\textwidth]{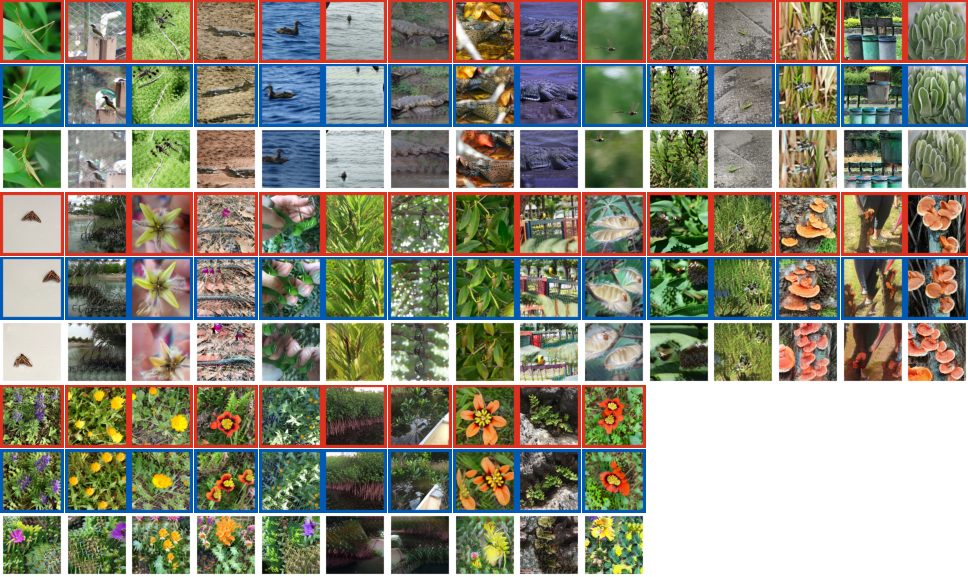}
    \caption{DINO on iNaturalist: Training Image (red), Inversion of Original Embedding (blue) and Inversion of Reconstructed Embedding.}
    \label{fig:gt_inversion_dino_inat}
\end{figure}

\begin{figure}
    \centering
    \includegraphics[width=\textwidth]{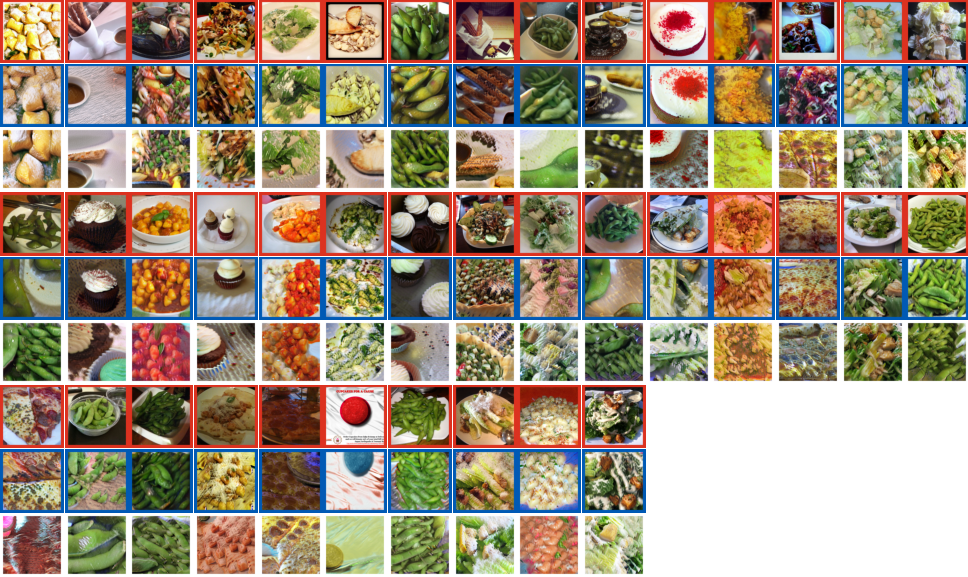}
    \caption{DINO2 on Food101: Training Image (red), Inversion of Original Embedding (blue) and Inversion of Reconstructed Embedding.}
    \label{fig:gt_inversion_dino2_food}
\end{figure}

\begin{figure}
    \centering
    \includegraphics[width=\textwidth]{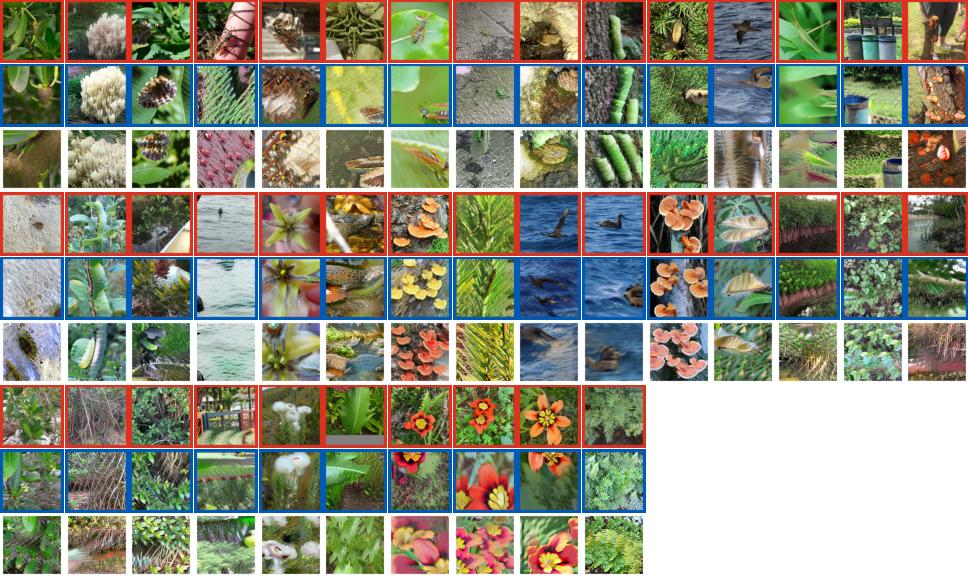}
    \caption{DINO2 on iNaturalist: Training Image (red), Inversion of Original Embedding (blue) and Inversion of Reconstructed Embedding.}
    \label{fig:gt_inversion_dino2_inat}
\end{figure}

\begin{figure}
    \centering
    \includegraphics[width=\textwidth]{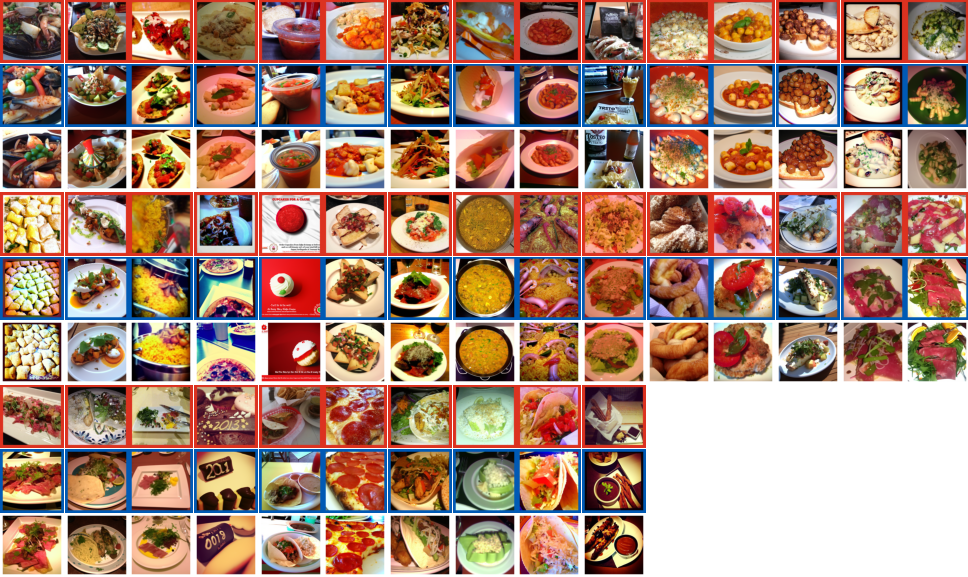}
    \caption{CLIP on Food101: Training Image (red), UnCLIP of Original Embedding (blue) and UnCLIP of Reconstructed Embedding.}
    \label{fig:gt_inversion_clip_food}
\end{figure}

\begin{figure}
    \centering
    \includegraphics[width=\textwidth]{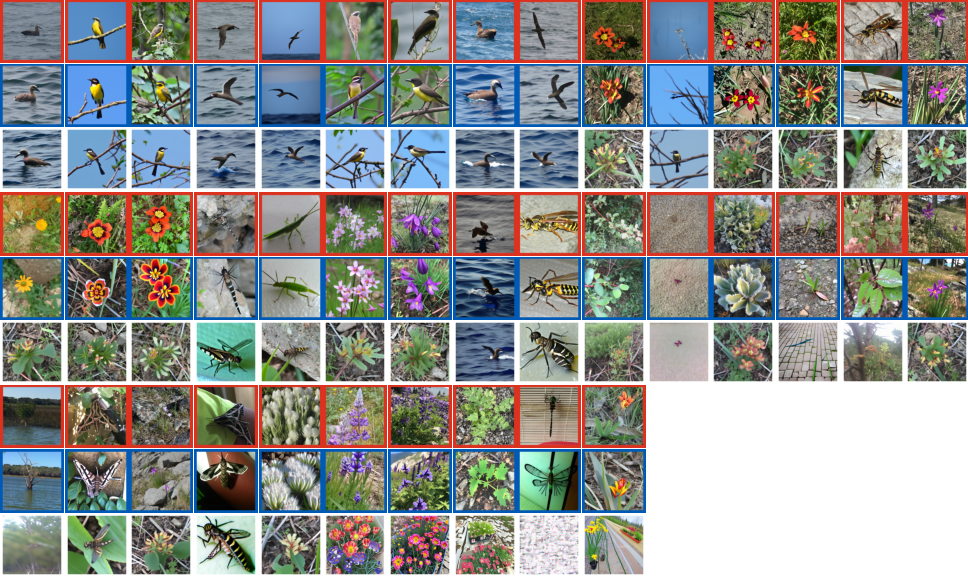}
    \caption{CLIP on iNaturalist: Training Image (red), UnCLIP of Original Embedding (blue) and UnCLIP of Reconstructed Embedding.}
    \label{fig:gt_inversion_clip_inat}
\end{figure}

\newpage
\section{Implementation Details}
\label{appen:implementation_details}

Our code is implemented with {\sc PyTorch}~\citep{paszke2019pytorch} framework. 

\subsection{Data Preprocessing}
\label{appen:preprocessing}

We resize each image to a resolution of $224$ pixels (the smaller side of the image) and then apply a center crop to obtain a $224 \times 224$ image. We then normalize the image per pixel following the normalization used in the original paper of each model, as shown in the table below:

\begin{center}
\begin{tabular}{|c|c|c|}
    \hline 
    Model & Mean & Std \\
    \hline 
    DiNO, DiNOv2 & $[0.485, 0.456, 0.406]$ & $[0.229, 0.224, 0.225]$ \\
    ViT & $[0.5, 0.5, 0.5]$ & $[0.5, 0.5, 0.5]$ \\
    CLIP & $[0.481, 0.458, 0.408]$ & $[0.269, 0.261, 0.276]$ \\
    \hline 
\end{tabular} \\
\end{center}

After feeding the images through the backbone $\F$ to obtain the image embeddings $\F(\bs_i)$, we normalize each embedding by subtracting the mean-embedding and dividing by the std. Formally: $\bx_i = \frac{\F(\bs_i)-\mu}{\sigma}$, where $\mu=\frac{1}{n}\sum_{i=1}^n\F(\bs_i)$, and $\sigma = \sqrt{\frac{1}{n-1} \sum_{i=1}^n \left( \F(\bs_i) -\mu \right)^2}$.

This is a fairly common approach when training on small datasets. $\mu$ and $\sigma$ can be thought as being part of the model $\phi$ as they are also applied for embeddings from outside the training set.

\subsection{Reconstruction Hyperparameter Search}
\label{appen:sweeps_details}

As mentioned in~\cref{sec:method_reconstruction}, we run the reconstruction optimization $100$ times with different choice of the $4$ hyperparameters of the reconstruction algorithm:
\begin{enumerate}
    \item Learning rate
    \item $\sigma$ -- the initial s.t.d. of the initialization of the candidates
    \item $\lambda_{\text{min}}$ -- together with the loss~\cref{eq:reconstruction_loss}, the reconstruction includes another loss term to require $\lambda_i>\lambda_{\text{min}}$ (a consequence of the KKT conditions is that $\lambda_i>0$, but if $\lambda_i=0$ it has no relevance in the overall results, therefore a minimal value $\lambda_{\text{min}}$ is set.).
    \item $\alpha$ -- Since the derivative of ReLU is piecewise constant and non-continuous, the backward function in each ReLU layer in the original model is replaced with the derivative of SoftRelu with parameter $\alpha$.
\end{enumerate}

For full explanation of the hyperparameters, please refer to ~\cite{haim2022reconstructing}. Note that for $m=500$, running $100$ times would result in $50$k candidates.

The hyperparameter search is done via Weights\&Biases~\citep{wandb}, with the following randomization (it is in the format of a W\&B sweep):

\begin{verbatim}
parameters:
  random_init_std:
    distribution: log_uniform_values
    max: 1 
    min: 1e-06
  optimizer_reconstructions.lr:
    distribution: log_uniform_values
    max: 1
    min: 1e-06
  loss.lambda_regularizer.min_lambda:
    distribution: uniform
    max: 0.5
    min: 0.01
  activation.alpha:
    distribution: uniform
    max: 500
    min: 10
\end{verbatim}

\subsection{Further Details about Inversion \cref{sec:method_inversion}}
\label{appen:imp_inversion}

We follow similar methodology to~\cite{tumanyan2022splicing}, using their code\footnote{\url{https://splice-vit.github.io/}} and changing the reconstruction loss from MSE to Cosine-Similarity as mentioned in~\cref{sec:method_inversion}, and specifically~\cref{eq:inversion_cossim} (see justifications in~\cref{appen:cossim_motivation}).

The Deep-Image Prior model $g$ is a fully convolutional U-Net model~\citep{ronneberger2015unet} (initialized at random with the default pytorch implementation). The optimization is run for $20$,$000$ iterations, where at each iteration the input to $g$ is $z + r$, where $z$ is initialized from $z \sim \mathcal{N}(\textbf{0}_{d_s}, \mathbb{I}_{d_s \times d_s})$ and kept fixed throughout the optimization, and $r$ is sampled at each iteration as follows:

\begin{tabular}{ll}
    iteration $i<10$,$000$: & $r \sim \mathcal{N}(\textbf{0}_{d_s}, 10\cdot \mathbb{I}_{d_s \times d_s})$ \\
    iteration $10$,$000 < i \leq 15$,$000$: & $r \sim \mathcal{N}(\textbf{0}_{d_s}, ~~2\cdot \mathbb{I}_{d_s \times d_s})$ \\
    iteration $15$,$000 < i \leq 20$,$000$: & $r \sim \mathcal{N}(\textbf{0}_{d_s}, 0.5\cdot \mathbb{I}_{d_s \times d_s})$
\end{tabular}

Note that the input to $g$ is of the same size of the input to $\F$, which is simply and image of dimensions $d_s = c \times h \times w$. At each iteration, the output of $g$ is fed to $\F$, and the output of $\F$ (which is an embedding vector of dimension $d=768$), is compared using cosine-similarity to the embedding vector that we want to invert. At the end of the step, the parameters of $g$ are changed to increase the cosine similarity between the embeddings.

\subsection{Inversion with UnCLIP}
\label{appen:unclip_inversion}

While the method in~\cref{appen:imp_inversion} is used for ViT, DINO and DINOv2, for CLIP we use a different method to invert, which is by using the UnCLIP implementation of~\cite{kakaobrain2022karlo-v1-alpha}. Unlike the inversion in~~\cref{appen:imp_inversion} that uses cosine-similarity, with UnCLIP, the embeddings (that go into to UnCLIP decoder) should have the right scale. For each CLIP embedding of a training image ($\bx$), we search for its nearest neighbour candidate ($\hat{\bx}$) with cosine similarity, but before feeding $\hat{\bx}$ into the UnCLIP decoder, we re-scale it to have the same scale as $\bx$, so that the input to the decoder is in fact $(\Vert \bx \Vert / \Vert \hat{\bx} \Vert) \hat{\bx}$. Unfortunately we could not resolve this reliance on the training set (as is also done in previous reconstruction works, and discussed in the main paper), but we believe this may be mitigated by computing and using general statistics of the training set (instead of specific training samples). We leave this direction for future research.

\subsection{Reconstruction in Multiclass Setup}
\label{appen:multiclass}

The method in~\cref{sec:method_reconstruction} was extended to multiclass settings by~\cite{buzaglo2023deconstructing}. In a nutshell, the reconstruction loss in~\cref{eq:reconstruction_loss} contains the gradient (w.r.t. $\btheta$) of $y_i \phi(\bx_i)$ which is the distance from the decision boundary. For multiclass model $\phi: \reals^d \to \reals^C$, the distance to the decision boundary is \mbox{$\phi(\bx_i)_{y_i} - \text{max}_{j \neq y_i} \phi(\bx_i)_{j}$ } . Replaced into the reconstruction loss in~\cref{eq:reconstruction_loss}, we have:

\[
L_{\text{rec}}(\hat{\bx}_1,\dots,\hat{\bx}_m,\lambda_1,\dots,\lambda_m):= \norm{\btheta - \sum_{i=1}^m\lambda_i \nabla_\btheta \left[ \phi(\hat{\bx}_i,  \btheta)_{y_i} - \underset{j \neq y_i}{\text{max}} \phi(\hat{\bx}_i, \btheta)_{j} \right]}_2^2 
\]

\subsection{Choice of Weight Decay}
\label{appen:wd}

\begin{figure}[htbp]
    \centering
    \includegraphics[width=.85\textwidth]{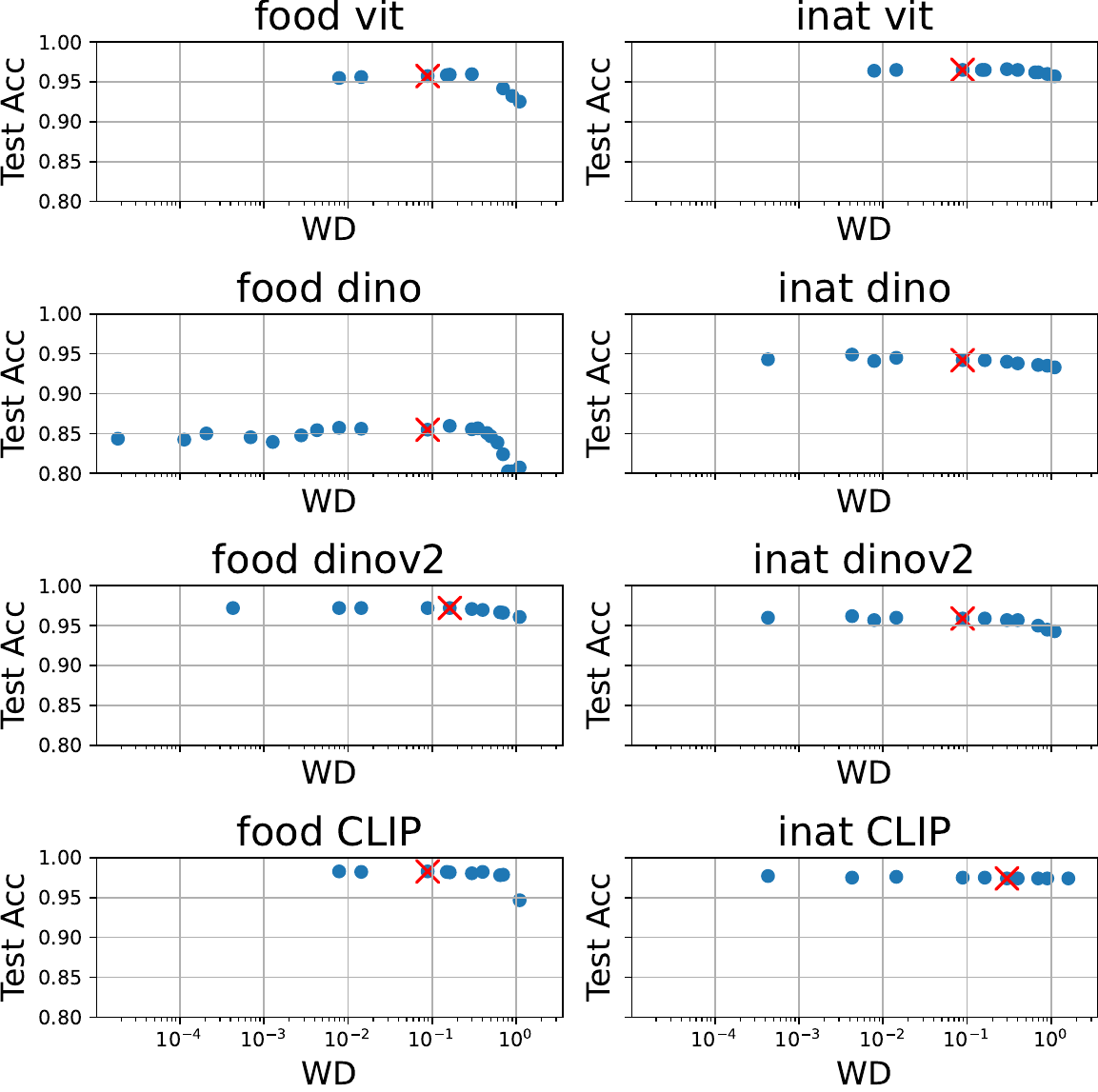}
    \caption{Test-Accuracy for different choices of Weight-Decay Value. Red-X marks the specific run used for reconstruction in~\cref{fig:main_results}}
    \label{fig:wd_selection}
\end{figure}

When training our model, we apply weight decay regularization. However, determining the optimal weight decay (WD) value is not straightforward. To find a WD value, we conduct a search across different WD values and observe their impact on test accuracy. The reuslts are shown in~\cref{fig:wd_selection}. We notice that for most values, the test accuracy increases until approximately 0.1 and then decreases from about 0.3 (indicating that WD is too large). We select the WD from this range, either 0.08 or 0.16. A red-x marks the run which was selected for reconstruction (and whose results are shown in~\cref{fig:main_results}).

\newpage
\section{Datasets - Full Details}
\label{appen:datasets}

\subsection{Image Resolution}

\begin{figure}[htbp]
    \centering
    \begin{tabular}{cc}
         \includegraphics[width=0.45\textwidth]{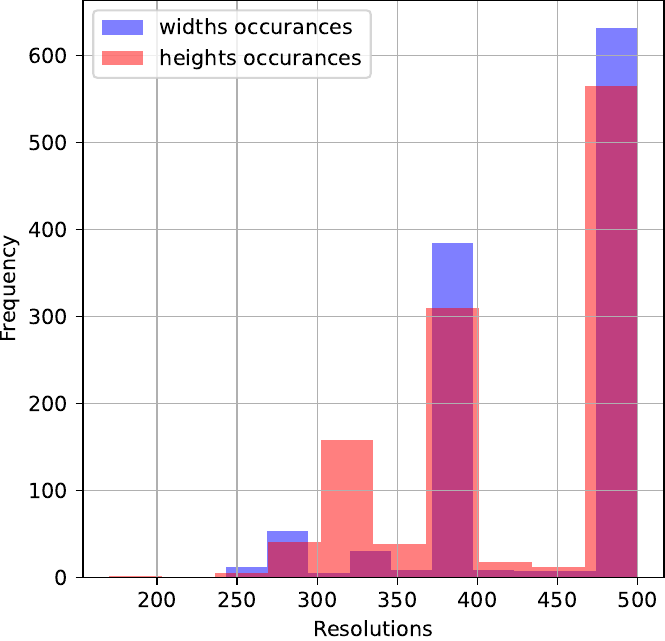}&
         \includegraphics[width=0.45\textwidth]{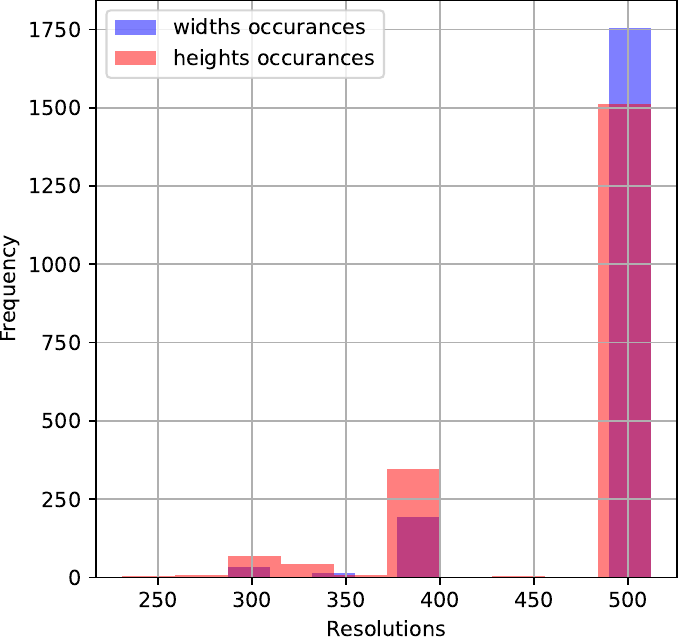} \\
         (a)&(b)
    \end{tabular}
    \caption{Resolution frequency of the images in use from iNaturalist (a) and Food101 (b) datasets}
    \label{fig:resolutions}
\end{figure}

\cref{fig:resolutions} illustrates how images in the datasets we used may have different resolutions. To standardize the input, we use the pre-processing described in~\cref{appen:preprocessing}.

\subsection{Food-101~\citep{bossard2014food}}

The dataset comprises real images of the 101 most popular dishes from the foodspotting website. 

\paragraph{Binary Tasks}
We use the following classes:
\begin{itemize}
    \item Class I:  "beef carpaccio", "bruschetta", "caesar salad", "churros" and "cup cakes"
    \item Class II:  "edamame", "gnocchi", "paella", "pizza" and "tacos"
\end{itemize}

For any choice of training samples amount, we randomly pick half from every such combined class in order to create our new dataset. 

\paragraph{Multiclass Tasks}
We use the following classes:

"beef carpaccio", "beet salad", "carrot cake", "cup cakes", "dumplings", "gnocchi", "guacamole", "nachos", "pizza" and "samosa"

Here the classes are not combined. For every choice of $N$ classes we choose the first $N$ out of the list above and randomly pick examples according to the training set size and in such a way that the newly formed dataset is balanced.

\subsection{iNaturalist~\citep{van2018inaturalist}}

The dataset encompasses a total of 10,000 classes, each representing a distinct species. 

\paragraph{Binary Tasks}

Classes are combined in the same manner as for the Food101 dataset. All classes names below appear as they are in the dataset.

\begin{adjustwidth}{2.5em}{0pt}

\textbf{Fauna}

\clstt{
02590\_Animalia\_Arthropoda\_Insecta\_Odonata\_Macromiidae\_Macromia\_taeniolata \\
02510\_Animalia\_Arthropoda\_Insecta\_Odonata\_Libellulidae\_Libellula\_forensis \\
02193\_Animalia\_Arthropoda\_Insecta\_Lepidoptera\_Sphingidae\_Eumorpha\_vitis \\
02194\_Animalia\_Arthropoda\_Insecta\_Lepidoptera\_Sphingidae\_Hemaris\_diffinis \\
00828\_Animalia\_Arthropoda\_Insecta\_Hymenoptera\_Vespidae\_Polistes\_chinensis \\
00617\_Animalia\_Arthropoda\_Insecta\_Hemiptera\_Pentatomidae\_Dolycoris\_baccarum \\
02597\_Animalia\_Arthropoda\_Insecta\_Orthoptera\_Acrididae\_Acrida\_cinerea \\
05361\_Animalia\_Mollusca\_Gastropoda\_Stylommatophora\_Philomycidae\_Megapallifera\_mutabilis \\
04863\_Animalia\_Chordata\_Reptilia\_Crocodylia\_Crocodylidae\_Crocodylus\_niloticus \\
04487\_Animalia\_Chordata\_Aves\_Procellariiformes\_Diomedeidae\_Phoebastria\_nigripes \\
04319\_Animalia\_Chordata\_Aves\_Passeriformes\_Tyrannidae\_Myiozetetes\_cayanensis \\}

\textbf{Flora}

\clstt{
05690\_Fungi\_Basidiomycota\_Agaricomycetes\_Polyporales\_Polyporaceae\_Trametes\_coccinea \\
05697\_Fungi\_Basidiomycota\_Agaricomycetes\_Russulales\_Auriscalpiaceae\_Artomyces\_pyxidatus \\
05982\_Plantae\_Tracheophyta\_Liliopsida\_Asparagales\_Iridaceae\_Olsynium\_douglasii \\
05988\_Plantae\_Tracheophyta\_Liliopsida\_Asparagales\_Iridaceae\_Sparaxis\_tricolor \\
06988\_Plantae\_Tracheophyta\_Magnoliopsida\_Asterales\_Asteraceae\_Silphium\_laciniatum \\
06665\_Plantae\_Tracheophyta\_Magnoliopsida\_Asterales\_Asteraceae\_Calendula\_arvensis \\
07032\_Plantae\_Tracheophyta\_Magnoliopsida\_Asterales\_Asteraceae\_Syncarpha\_vestita \\
07999\_Plantae\_Tracheophyta\_Magnoliopsida\_Fabales\_Fabaceae\_Lupinus\_arcticus \\
07863\_Plantae\_Tracheophyta\_Magnoliopsida\_Ericales\_Primulaceae\_Myrsine\_australis \\
08855\_Plantae\_Tracheophyta\_Magnoliopsida\_Malpighiales\_Rhizophoraceae\_Rhizophora\_mangle \\
09143\_Plantae\_Tracheophyta\_Magnoliopsida\_Ranunculales\_Berberidaceae\_Berberis\_bealei \\
09974\_Plantae\_Tracheophyta\_Polypodiopsida\_Polypodiales\_Pteridaceae\_Cryptogramma\_acrostichoides \\
}

\end{adjustwidth}

\paragraph{Multiclass Tasks}

\begin{adjustwidth}{2.5em}{0pt}

\textbf{1. Insects}

\clstt{
02590\_Animalia\_Arthropoda\_Insecta\_Odonata\_Macromiidae\_Macromia\_taeniolata \\
01947\_Animalia\_Arthropoda\_Insecta\_Lepidoptera\_Nymphalidae\_Phaedyma\_columella \\
02194\_Animalia\_Arthropoda\_Insecta\_Lepidoptera\_Sphingidae\_Hemaris\_diffinis \\
02195\_Animalia\_Arthropoda\_Insecta\_Lepidoptera\_Sphingidae\_Hemaris\_fuciformis \\
02101\_Animalia\_Arthropoda\_Insecta\_Lepidoptera\_Pieridae\_Pontia\_occidentalis \\
02138\_Animalia\_Arthropoda\_Insecta\_Lepidoptera\_Riodinidae\_Apodemia\_virgulti \\
}\\

\textbf{2. Aquatic Animals}

\clstt{
02715\_Animalia\_Arthropoda\_Malacostraca\_Decapoda\_Grapsidae\_Grapsus\_grapsus \\
02850\_Animalia\_Chordata\_Actinopterygii\_Perciformes\_Lutjanidae\_Ocyurus\_chrysurus \\
02799\_Animalia\_Chordata\_Actinopterygii\_Perciformes\_Centrarchidae\_Ambloplites\_rupestris \\
02755\_Animalia\_Arthropoda\_Merostomata\_Xiphosurida\_Limulidae\_Limulus\_polyphemus \\
02704\_Animalia\_Arthropoda\_Malacostraca\_Decapoda\_Cancridae\_Cancer\_borealis \\
02706\_Animalia\_Arthropoda\_Malacostraca\_Decapoda\_Cancridae\_Cancer\_productus \\
}\\

\textbf{3. Reptiles}

\clstt{
04859\_Animalia\_Chordata\_Reptilia\_Crocodylia\_Alligatoridae\_Alligator\_mississippiensis \\
04868\_Animalia\_Chordata\_Reptilia\_Squamata\_Agamidae\_Agama\_picticauda \\
04862\_Animalia\_Chordata\_Reptilia\_Crocodylia\_Crocodylidae\_Crocodylus\_moreletii \\
04865\_Animalia\_Chordata\_Reptilia\_Rhynchocephalia\_Sphenodontidae\_Sphenodon\_punctatus \\
04954\_Animalia\_Chordata\_Reptilia\_Squamata\_Colubridae\_Pituophis\_deppei \\
}\\

\textbf{4. Birds}

\clstt{
04487\_Animalia\_Chordata\_Aves\_Procellariiformes\_Diomedeidae\_Phoebastria\_nigripes \\
04319\_Animalia\_Chordata\_Aves\_Passeriformes\_Tyrannidae\_Myiozetetes\_cayanensis \\
04570\_Animalia\_Chordata\_Aves\_Suliformes\_Phalacrocoracidae\_Microcarbo\_melanoleucos \\
04587\_Animalia\_Chordata\_Aves\_Suliformes\_Sulidae\_Sula\_nebouxii \\
04561\_Animalia\_Chordata\_Aves\_Strigiformes\_Strigidae\_Surnia\_ulula \\
04576\_Animalia\_Chordata\_Aves\_Suliformes\_Phalacrocoracidae\_Phalacrocorax\_capensis \\
}
\end{adjustwidth}


\end{document}